\documentclass{article} %
\usepackage{iclr2025_conference,times}

\usepackage{amsmath,amsfonts,bm}

\def\eqref#1{equation~\ref{#1}}

\def\1{\bm{1}}

\def\ra{{\textnormal{a}}}

\DeclareMathAlphabet{\mathsfit}{\encodingdefault}{\sfdefault}{m}{sl}
\SetMathAlphabet{\mathsfit}{bold}{\encodingdefault}{\sfdefault}{bx}{n}

\DeclareMathOperator*{\argmax}{arg\,max}

\usepackage{hyperref}
\usepackage{url}
\usepackage{graphicx}

\usepackage{float}
\floatstyle{plaintop}
\restylefloat{table}

\usepackage{algorithm}
\usepackage[noend]{algpseudocode}

\usepackage{tabularx}

\newcommand{\gapxx}{\hspace*{4mm}}

\usepackage[compact]{titlesec} 
\titlespacing\section{0pt}{2pt plus 0pt minus 1pt}{1.5pt plus 0pt minus 1pt}
\titlespacing\subsection{0pt}{2pt plus 0pt minus 1pt}{1.5pt plus 0pt minus 1pt}
\makeatletter
\renewcommand{\paragraph}{%
  \@startsection{paragraph}{4}%
  {\z@}{0.05ex \@plus .05ex \@minus .05ex}{-1em}%
  {\normalfont\normalsize\bfseries}%
}
\usepackage[font={small}]{caption}

\newcommand{\eat}[1]{}
\usepackage{listings}
\lstdefinestyle{base}{
  emptylines=1,
  breaklines=true,
  linewidth=\textwidth,
  commentstyle=\color{gray},
  basicstyle=\scriptsize\ttfamily,
  moredelim=**[is][\color{blue}]{@}{@},
}
\usepackage{booktabs}
\usepackage{amsmath}
\usepackage{mathtools}

\newcommand\scalemath[2]{\scalebox{#1}{\mbox{\ensuremath{\displaystyle #2}}}}

\usepackage{enumitem}

\newcommand{\highlighthtml}[2]{\colorbox[HTML]{#1}{#2}}

\newcommand{\treewise}{\textsc{TreeWise}}
\usepackage{colortbl}
\definecolor{lightgreen}{RGB}{223,255,219}
\definecolor{lightred}{RGB}{255,219,219}
\definecolor{lightblue}{RGB}{219,235,255}
\definecolor{lightorange}{RGB}{255,235,219}
  
\newcommand{\up}{\cellcolor{lightblue}}  
\newcommand{\dn}{\cellcolor{lightorange}} 

\definecolor{lightgreen}{RGB}{223,255,219}
\definecolor{lightred}{RGB}{255,219,219}
\definecolor{redish}{RGB}{255,102,102}
\definecolor{blueish}{RGB}{31, 78, 192}
\definecolor{orangeish}{RGB}{255, 178, 102}
\definecolor{mypink1}{rgb}{0.858, 0.188, 0.478}
\definecolor{mypink2}{RGB}{219, 48, 122}
\definecolor{mypink3}{cmyk}{0, 0.7808, 0.4429, 0.1412}
\definecolor{mygray}{RGB}{220, 220, 220}
\definecolor{darkbluee}{RGB}{0,17, 113}
\definecolor{purpleNew}{RGB}{151, 45, 204}
\definecolor{purplebg}{RGB}{229, 199, 244}
\definecolor{violet}{rgb}{0.70,0.05,0.65}

\usepackage{adjustbox}
\usepackage{comment}
\renewcommand{\cite}{\citep}
\usepackage{subcaption}  %
\newcommand{\myparagraph}[1]{\noindent\textbf{#1}}
\newcommand{\sref}[1]{\S\ref{#1}}
\usepackage{wrapfig}

\title{From Models to Microtheories: \\ 
Distilling a Model's Topical Knowledge \\
for Grounded Question Answering 
}

\author{Nathaniel Weir\textsuperscript{\rm \textdagger\thanks{Work done in part during internship at Allen Institute for AI}},
    Bhavana Dalvi Mishra\textsuperscript{\rm $\spadesuit$},
    Orion Weller\textsuperscript{\rm \textdagger},  
    Oyvind Tafjord\textsuperscript{\rm $\spadesuit$},\\
    \textbf{Samuel Hornstein\textsuperscript{\rm $\#$},
    Alexander Sabol\textsuperscript{\rm $\#$},
    Peter Jansen\textsuperscript{\rm $\spadesuit$}\textsuperscript{\rm $\ddagger$}, 
    }\\
\textbf{Benjamin Van Durme\textsuperscript{\rm \textdagger},  
    Peter Clark\textsuperscript{\rm $\spadesuit$}} \\
      $^{\text{\textdagger}}$Johns Hopkins University \quad 
  $^{\spadesuit}$Allen Institute for AI\\
  $^{\#}$Thomas Jefferson University \quad
  $^{\ddagger}$University of Arizona \\
  \texttt{nweir@jhu.edu}, 
 \texttt{peterc@allenai.org} \\
}

\iclrfinalcopy %
\begin{document}

\maketitle

\begin{abstract}

Recent reasoning methods (e.g., chain-of-thought, 
entailment reasoning)
help users understand how language models (LMs) answer a single question, but
they do little to reveal the LM's {\it overall understanding},
or ``theory", about the question's {\it topic}, making it 
still 
hard to trust the model.
Our goal is to materialize such theories - here called {\it microtheories} (a linguistic
analog of logical microtheories~\cite{blair-etal-1992-microtheories}) - as
a set of sentences encapsulating an LM's core knowledge about a topic.
These statements
systematically work together to entail answers to a {\it set} of questions
to both engender trust and improve performance.
Our approach is to first populate a knowledge store with (model-generated)
sentences that entail answers to training questions, and then
distill those down to a core microtheory which is concise, general,  
and non-redundant.
We show that, when added to a general corpus (e.g., Wikipedia), microtheories
can supply critical, topical information not necessarily present in the corpus,
improving both a model's ability to ground its answers 
to verifiable knowledge
(i.e., show how answers are systematically entailed by
documents in the corpus, fully grounding up to +8\% more answers), and the accuracy of those
grounded answers (up to +8\% absolute).
We also show that, in a human evaluation in the medical domain, 
our distilled microtheories contain a significantly higher
concentration of topically critical facts than the non-distilled
knowledge store. Finally, we show we can quantify the coverage of a
microtheory for a topic (characterized by a dataset) using a notion of {\it $p$-relevance}. 
Together, these suggest that microtheories are an efficient
distillation of an LM’s topic-relevant knowledge, 
that they can usefully augment existing corpora, 
and can provide both performance gains and an interpretable, verifiable window into the model’s knowledge of a topic.\footnote{All code and data for this work 
will be released at \url{https://github.com/nweir127/microtheories/}.}

\end{abstract}

\section{Introduction}
What do language models (LMs) know about the topics they converse about? Despite
their success, LMs are opaque - users do not have good visibility into
the LM's knowledge, creating challenges for engendering trust in the model.
Recent work has tried to address this for answers to a single question
by having the LM provide various explanations and chains of thought
that support an answer, e.g., \cite{danilevsky-etal-2020-survey,wei-etal-2022-chain}. However, such
expositions only reveal snippets of the LM's latent knowledge and do little to
convey the model's overall understanding, or ``theory'', about the topic at hand,
limiting a user's confidence in the model's behavior. Conversely, if we had a method
that could distill a model's overall theories into an inspectable, verifiable form,
this could help users see how answers follow from those theories.

\begin{figure}
    \centering
    \includegraphics[trim={7mm 3mm 6mm 3mm},clip,width=0.65\linewidth]{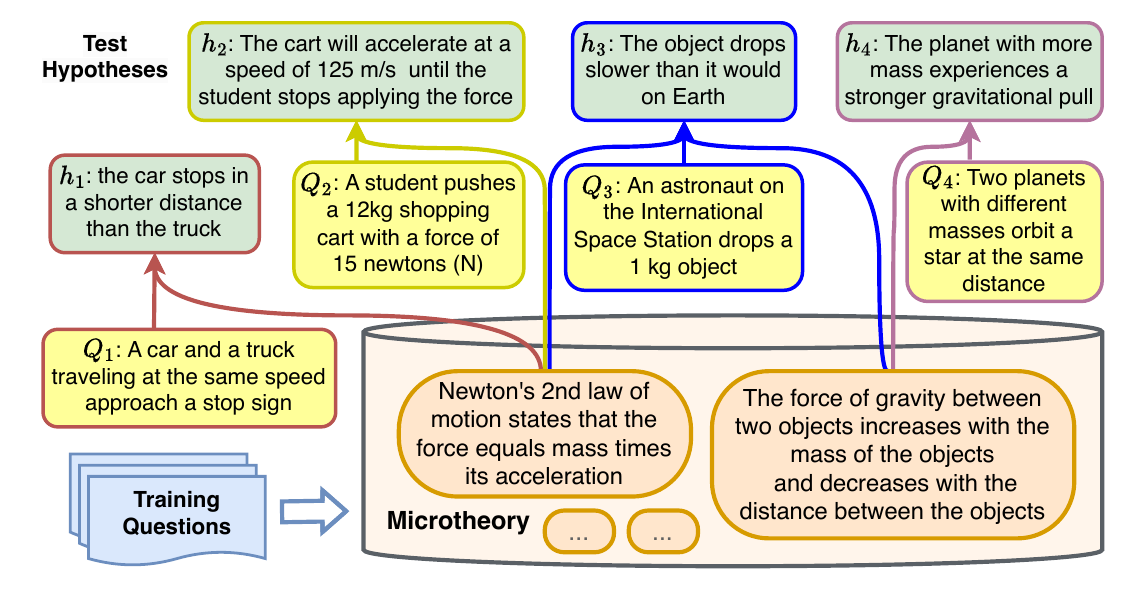}
\caption{Given a set of topical training questions, we construct a microtheory, a set of statements
articulating a model's core, reusable knowledge about that topic. These help
prove (entail) answers to test questions.}
    \label{fig:mt-cartoon}
\end{figure}

Our goal is to distill such theories - which we refer to as {\it microtheories} (Mts)
(a linguistic analog of logical microtheories~\cite{blair-etal-1992-microtheories}) - as
a set of sentences encapsulating an LM's core knowledge about a {\it topic}.
Our approach uses a question-driven methodology, where a set of questions characterizes a
topic (i.e., a topic is defined as a distribution over questions).
We first prompt the LM to enumerate facts that entail answers to training questions,
then identify the core, abstractable pieces of knowledge frequently reused for many questions
in that set, then distill them down to a core microtheory that is concise, 
general,
and non-redundant. The microtheory is intended to contain the core principles about the
topic (e.g., Newton's laws of motion, for physics), extracted from the LM, that can help derive corpus-grounded answers to topical questions.

To answer questions with the help of a microtheory, we use the well-known
process of {\it textual entailment} \cite{dagan-etal-2005-pascal, lai-etal-2017-natural}. Textual entailment is a linguistic
inference process that identifies when one (or more) statement(s) ``reasonably
implies'' another \cite{magnini-2015-book}, producing a (potentially multistep) entailment
tree, analogous to a proof tree~\cite{dalvi-etal-2021-explaining}. 
While entailment lacks
a fully formal definition (modern entailment engines are instead trained on examples),
it provides a well-defined mechanism for deciding if, and how, an answer systematically
follows from a set of statements.
This allows us to both use and evaluate microtheories for question-answering.

\eat{
Of course, reasoning about broad domains requires not just core principles,
but also an endless number of idiosyncratic facts that are unlikely to be contained
in the microtheory. For example, determining that a dropped pencil (say) will
fall to the ground requires not just general knowledge of gravity, but also the
idiosyncratic fact that a pencil is a denser-than-air object. To supply
such facts, we assess our microtheories when combined with a large, general
corpus (Wikipedia).
}

We evaluate our microtheory approach in two domains using two existing
datasets: {\bf K-12 science}, using the ARC dataset \cite{clark-etal-2018-think},
and {\bf medicine}, using the MedQA dataset \cite{jin-etal-2021-disease}.
We partition each dataset into topics (via clustering), then construct
microtheories on a training split of each topic, and test on the remainder.
We find that when added to a general corpus (here, Wikipedia), microtheories
can supply critical, topical information not necessarily present in the corpus,
improving both the model's ability to ground its answers 
(i.e., show how those answers are entailed by the corpus, fully grounding up to +8\% more answers),
and the accuracy of those grounded answers (up to +8\% absolute).
We also show that, in a human evaluation in the medical domain, 
our distilled microtheories contain a significantly higher
concentration of topically critical facts than in the non-distilled
knowledge store. 
Finally, we propose a metric, {\it $p$-relevance}, to quantify the completeness of a microtheory.
By correlating {$p$-relevance with training set size, we can predict how many
training questions will be needed to generate a Mt relevant to some $p$ percent
of test questions.
In summary, our contributions are:
\begin{itemize}[topsep=0pt,itemsep=0ex,partopsep=1ex,parsep=1ex,labelindent=0pt,wide,labelwidth=!]
\item A method for distilling microtheories from a model, given a QA dataset, that
  articulates the model's core knowledge about the dataset topic. This is the first method
  to try to distill a model's overall understanding of a topic in this way.
\item Evaluations that show microtheories can improve both a model's ability to
  ground its answers to verifiable facts in a corpus, and the accuracy of those answers.
\item A human evaluation in the medical domain, showing our method 
  distills out topically critical facts from the general pool.
\item A new metric, {\it $p$-relevance}, to quantify the completeness of a microtheory for a dataset.
\end{itemize}
Together, these suggest that microtheries are an efficient distillation of an LM's topic-relevant knowledge, 
that they can usefully augment existing corpora, and can provide both performance gains and an interpretable,
verifiable window into the model's knowledge of a topic.

\section{Related Work}

The notion of a microtheory - a small, coherent, topical representation of knowledge -
arose in formal logic to help organize a large, logical KB into useful,
topical subsets, each sharing the same underlying assumptions (context) \cite{lenat-etal-1990-cyc,blair-etal-1992-microtheories}.
A logical microtheory aimed to model a specific aspect of reality as a
self-contained, logical system, akin to a piece of ``computational clockwork'' \cite{clark-etal-2004-knowledge}.
Our work here can be seen as a modern approach to this, similarly
aiming to capture an understanding of a topic but where the units of knowledge
are in NL and reasoning is via textual entailment.

\eat{
It has long been recognized that LMs contain substantial knowledge, and can even be viewed
as a new kind of ``knowledge base'' \cite{petroni-etal-2019-language,bosselut-etal-2019-comet,weir-etal-2020-probing,Alivanistos2022PromptingAP}. Recent work has sought
to systematically materialize some of this knowledge as natural language
statements via self-querying, e.g., building a BeliefBank \cite{kassner-etal-2021-beliefbank},
or Belief Graphs \cite{kassner-etal-2023-language,tafjord-etal-2022-entailer,jung-etal-2022-maieutic} showing
which facts are ``believed'' by the model and how those facts support an answer. However, while providing
some visibility, these methods only provide disconnected slices of the model's
latent knowledge related to a {\it single question}. Our work takes this
further, aiming to extract a model's overall understanding about a {\it topic}.
}
It has long been recognized that LMs contain substantial knowledge, and can even be viewed
as a new kind of ``knowledge base'' \cite{petroni-etal-2019-language,bosselut-etal-2019-comet}. %
Some work has sought to materialize some of the model's factual knowledge either as
knowledge graph triples \cite{Alivanistos2022PromptingAP,Hao2022BertNetHK}, or
NL statements, e.g., BeliefBank \cite{kassner-etal-2021-beliefbank}.
Other work has generated ``belief graphs'' showing how model-believed facts entail answers
\cite{kassner-etal-2023-language,tafjord-etal-2022-entailer,jung-etal-2022-maieutic}.
However, while providing some visibility, these methods only provide disconnected slices of
the model's latent knowledge, at best related to a {\it single question}. Our work takes this
further, aiming to extract a model's overall understanding of a {\it topic}.

One notable manual attempt at a topical, linguistic microtheory is WorldTree,
a corpus of several thousand statements (aiming to) cover all the knowledge
required for a syllabus in elementary science \cite{xie-etal-2020-worldtree}, constructed
at significant expense. Our goal with microtheories can be seen as automating such a
construction process, and extracting such topical knowledge from language models directly (we later compare our Mts against WorldTree in \S\ref{sec:test_grounding} and \S\ref{sec:mt_qa_results}).

More broadly, there has been substantial work in generating explanations with language models,
to help users 
see the rationale behind a model's answers \cite{wiegreffe-marasovic-2021-teach,danilevsky-etal-2020-survey,wei-etal-2022-chain}.
However, much of this work is again question-centric, and sometimes lacks a clear
definition of whether an explanation adequately ``supports'' an answer.
To more clearly define the relationship between supporting evidence and an answer,
and for linguistic reasoning in general, there has been a resurgence of
interest in {\it textual entailment} as a linguistic inference process,
with numerous, high-quality, multi-step entailment engines becoming
available,  
e.g., SCSearch \cite{scsearch}, NLProofS \cite{NLProofS}, Entailer \cite{tafjord-etal-2022-entailer}, IRGR \cite{ribeiro-etal-2022-entailment},
TreeWise \cite{weir-etal-2024-enhancing}, and MetGen \cite{metgen}.
In our work, we leverage this development to
clearly define whether a microtheory supports an answer or not.

Finally, our work is distinct from work on library learning, which seeks to induce generalizations over data,
e.g., \cite{muggleton-etal-1991-inductive,Wang2023VoyagerAO}.
Rather, our goal here is to articulate generalizations that the LM appears to have
already formed.
\section{Approach}

\subsection{Background: Textual Entailment}

To make meaningful claims about microtheories, we need a clear definition of what
it means for an answer to follow from a microtheory (or a corpus in general). For this,
we use the well-known notion of {\it textual entailment} in NLP: a linguistic inference
process that identifies whether one or more statements ``reasonably imply'' another~\cite{dagan-etal-2005-pascal}.
Modern entailment engines perform a soft version of theorem proving by searching
for {\it entailment trees}, analogous to proof trees, showing how a query hypothesis $H$ follows
from statements in a theory $\mathcal{T}$, where both elements are stated in NL.
Here, H is a sentence (commonly a declarative answer to a question $Q$),
$\mathcal{T} = [t_1, \dots t_n]$ is a corpus of text passages, e.g. sentences or documents,
and an entailment tree $T$ is a tree whose root is $H$, whose leaves $L = [l_1, \dots, l_l]$
are a subset of texts in $T$, 
and whose individual steps have the form
$[p_1, \dots p_j] \models_{Q} h_k; \ j \in \{1, \dots,  |T|-1\}$, where $\models_Q$ denotes
atomic textual entailment given the context of question $Q$ ($\models_Q$ is
typically learned from examples).
Multiple alternative trees $T_i$ may be found supporting the
same hypothesis $H$, analogous to finding multiple proofs for a hypothesis in formal logic.
For large $\mathcal{T}$s, the search space can be very big. To handle this, entailment engines such
as IRGR~\cite{ribeiro-etal-2022-entailment} or \textsc{TreeWise}~\cite{weir-etal-2024-enhancing}
use iterative text retrieval to look up relevant statements and use them for grounding and
decomposing hypotheses.

{\bf Definitions:} If a tree $T_i$ shows how $H$ is entailed from a set of corpus texts $L_i$, we
say $H$ is {\bf grounded} in the corpus. We also say the leaf statements $L_i$ provide
an {\bf argumentative basis} for accepting $H$. If there are multiple trees with
different leafsets, we say that $L_1, L_2, \dots$ constitute {\bf alternative bases}
for supporting $H$.

\subsection{Constructing Microtheories}
\label{sec:methods}

We now describe a method to automatically construct a microtheory, namely a
list of model-generated NL assertions that are ``most important'' for entailing
answers to questions about a given topic, characterized by a set of training questions.
Microtheories are intended to capture the core knowledge about the topic,
and we evaluate different operationalizations of ``most important'' shortly.
First, for a given training set $[Q_1,\dots Q_d]$, we extract from an LM, via prompting, sets of facts that it would use to support the correct answer to each question (\S\ref{sec:extraction}).
Then, to reduce redundancies and promote generalizability, we apply a series of optimizations to distill the core kernels of knowledge by identfying entailment relations (\S\ref{sec:entailment_condense}).
Finally, we propose (and later evaluate) three alternative ways of 
optimizing the microtheory contents given a
particular size budget (\S\ref{sec:usage_optimization}).

\eat{
We construct a method to automatically populate a list of
NL assertions for a given task dataset that an entailment engine most commonly uses to answer questions. 
We propose to use an LLM, in conjunction with an entailment engine, to dynamically instantiate the assertions.
For a given dataset $[Q_1,\dots Q_d]$, we extract from an LM, via prompting, sets of facts that it would use to support the correct answer to each question (\S\ref{sec:extraction}).
To reduce redundancies and promote generalizability, we apply a series of optimizations to distill the core kernels of knowledge by identfying entailment relations (\S\ref{sec:entailment_condense}) and optimizing for fact usage by the engine (\S\ref{sec:usage_optimization}).
}

\subsubsection{Raw Fact Pool Extraction}
\label{sec:extraction}

We first prompt an LM to regurgitate facts on a question-by-question basis in order to populate an initial factpool $\mathcal{F}$. 
We use a few-shot learning prompt that cues the LM to perform a version of `chain-of-thought' reasoning~\cite{wei-etal-2022-chain} catered towards extracting facts. Each example in the prompt 
contains a question $Q_e$, answer options to $Q_e$ in the form of hypotheses $h_1, \dots h_n$, a list of supporting statements for $Q_e$ (a question-specific `theory'), and a sequence of entailment steps that compose a subset of statements into complex inferences that answer the question by supporting the correct $h$. Prompting details can be found in \S\ref{app:theorycot}. 
As training examples, we use EntailmentBank~\cite{dalvi-etal-2021-explaining}, a dataset of science questions paired with lists of supporting science facts that combine into an entailment tree for the correct answers.
\begin{align}
\scalemath{.87}{
\mathcal{F} = \bigcup_1^{d} \text{KeepIfGeneric}(\text{GenFacts}(Q_i, [h_1, \dots, h_n]))
}
\end{align}
Because the LLM often generates context-specific sentences via GenFacts$()$ that we wouldn't want to keep in our microtheory, e.g. about specific entities in the question (e.g. ``the man is running''),
we create a filter (KeepIfGeneric$()$ in (1)) that prompts the LLM to classify each sentence as generic or context-specific and keeps only the generics.

\subsubsection{SBERT and Entailment-based Condensation}
\label{sec:entailment_condense}
As fact lists are extracted independently for each question, 
there is substantial redundancy within the unified pool.
We use a pair of techniques to remove statements and 
produce condensed pool $\mathcal{C} \subseteq \mathcal{F}$:

\myparagraph{Soft Deduplication.} We identify paraphrases using a Sentence Transformer~\cite[SBERT;][]{reimers-gurevych-2019-sentence} trained to maximize the embedding cosine similarity between same-meaning facts.\footnote{\href{https://huggingface.co/sentence-transformers/all-mpnet-base-v2}{\texttt{all-mpnet-base-v2}}}
We perform a linear pass of $\mathcal{F}$ and remove every fact $f_i$ for which $\exists f_j \in \mathcal{F}, j > i, \cos(\text{SBERT}(f_i), \text{SBERT}(f_j)) > t$ for some threshold $t$.

\myparagraph{Entailment Condensation.}
We perform a second pass through $\mathcal{F}$ to find pairs $f_i, f_j$ s.t. $f_i \models f_j$, and remove $f_j$. We use a fine-tuned neural cross-encoder to predict $\models$.\footnote{\href{https://huggingface.co/sileod/deberta-v3-large-tasksource-nli}{\texttt{sileod/deberta-v3-large-tasksource-nli}}}
As it would be infeasible to check all $|\mathcal{F}|^2$ pairs of facts, we leverage SBERT to find the most promising candidates. We check for entailment between any two facts $f_i, f_j$ iff $\cos(\text{SBERT}(f_i), \text{SBERT}(f_j)) > u, u < t$.\footnote{We use $t=.9$ and $u = .3$.}

\subsubsection{Engine Usage-based Optimization}
\label{sec:usage_optimization}
After removing duplicates and entailments, we still see facts that serve the same \textit{functional} roles in many questions but imply slightly different things, e.g. for physics:
\begin{enumerate}[label=(\Alph*)]
  \setlength\itemsep{0em}
    \item The force required to cause a given acceleration is determined by the mass of an object
    \item The force acting on an object equals mass times acceleration
\end{enumerate}

While this might be acceptable for a knowledge store with infinite storage, we instead seek a more concise representation without redundancies.
Towards this goal, we consider three alternative approaches, illustrated in \autoref{fig:comp_approaches}, to identify some $n$ most effective statements to retain in the Mt. We will refer to the resulting sets as
{\bf $n$-Mts}.\footnote{$n$
is a hyperparameter; below, we observe that $n$ represents a trade-off between concision and performance.} Under this budget, we would not want to retain both (A) and (B) in the $n$-Mt if it meant discarding some (C) that is important to other questions.

\myparagraph{1. Top-$n$ Most Used Facts (Mt$_\text{usage}$).}
We seek the set of facts that best ``covers'' the argumentative bases for a given dataset of hypotheses. One intuitive way to choose these facts is to feed them to an entailment engine and see which ones are used most frequently to explain hypotheses in the dataset. We refer to this approach as the \textit{usage} $n$-Mt. Using questions $Q_1 \dots Q_d$ and hypotheses $h_1 \dots h_d$:
\begin{align}
\scalemath{.87}{
n\text{-Mt}_\text{usage} = \argmax_{\mathcal{M} \subseteq \mathcal{C}, |\mathcal{M}| = n} \sum_{f_i \in \mathcal{M}} \sum_{j=1}^{d} \delta{}(f_i \in  \textsc{leaves}(\textsc{engine}(h_j, \mathcal{C}; Q_j)))
}
\end{align}
where \textsc{engine} returns a (possibly empty) set of entailment trees proving $h_j$ from 
fact pool $\mathcal{C}$, and \textsc{leaves} refers to the set of all facts used at least once 
in any of those trees.
This objective does \textit{not} address the functional redundancy 
issue for fact pairs such as (A) and (B). If (A) and (B) serve the same role in entailing hypotheses, 
then both facts will be used for a similar number of questions; if one is in the top-$n$ most used facts, the other will likely also be. 

\myparagraph{2. Maximum Question Coverage (Mt$_\text{QC}$).}
We use an objective function minimizing the facts in a theory while maximizing the number of train hypotheses ``covered.'' For each question $Q_j$ with bases $\mathcal{L}_j = L_{1j}, L_{2j}, \dots$ (returned by $\textsc{LeafSets}$),  we try to fully cover at least one $L_{kj}$ for as many questions as possible using $\leq n$ facts. We use a linear program described in \S\ref{app:qc_ilp}.
\begin{align}
\begin{aligned}
\scalemath{.87}{n\text{-Mt}_\text{QC} = } & \scalemath{.87}{\argmax_{\mathcal{M} \subseteq \mathcal{C}, |\mathcal{M}| = n} \sum_{j=1}^{d} 
 \delta{}(\exists L_{kj} \in  \textsc{LeafSets}(\textsc{engine}(h_j, \mathcal{C}; Q_j)); L_{kj} \subseteq \mathcal{M})
}
\end{aligned}
\end{align}

\myparagraph{3. Maximum Partial Coverage (Mt$_\text{PC}$).}
\label{sec:partial_coverage}
Equation (3) poses question coverage as a binary variable: for each $Q_i$, either an entire basis is kept, or $Q_i$ is not covered. We consider a variant that loosens this constraint, modeling the \textit{partial} coverage of questions.\footnote{For 3 questions, it might be better to cover 70\%, 70\%, \& 70\% of a basis for each than 100\%, 100\%, \& 0\%. } 
We use an objective function that maximizes the per-question partial basis coverage (LP described in \S\ref{app:pc_ilp}). 
Given each question $Q_i$ associated with bases $L_{1i}, L_{2i}, \dots$, we try to best cover some $L^{*}_{ki}$. 
\begin{align}
\begin{aligned}
\scalemath{.87}{n\text{-Mt}_\text{PC} = }  \scalemath{.87}{\argmax_{\mathcal{M} \subseteq \mathcal{C}, |\mathcal{M}| = n} \sum_{j=1}^{d} } \scalemath{.87}{
 \argmax_{L_{kj} \in  \textsc{LeafSets}(\textsc{engine}(h_j, \mathcal{C}; Q_j))}
 \frac{|\mathcal{M} \cap L_{kj}|}{|L_{kj}|}
}
\end{aligned}
\end{align}

\begin{figure}[t!]
    \centering
    \includegraphics[width=.75\textwidth]{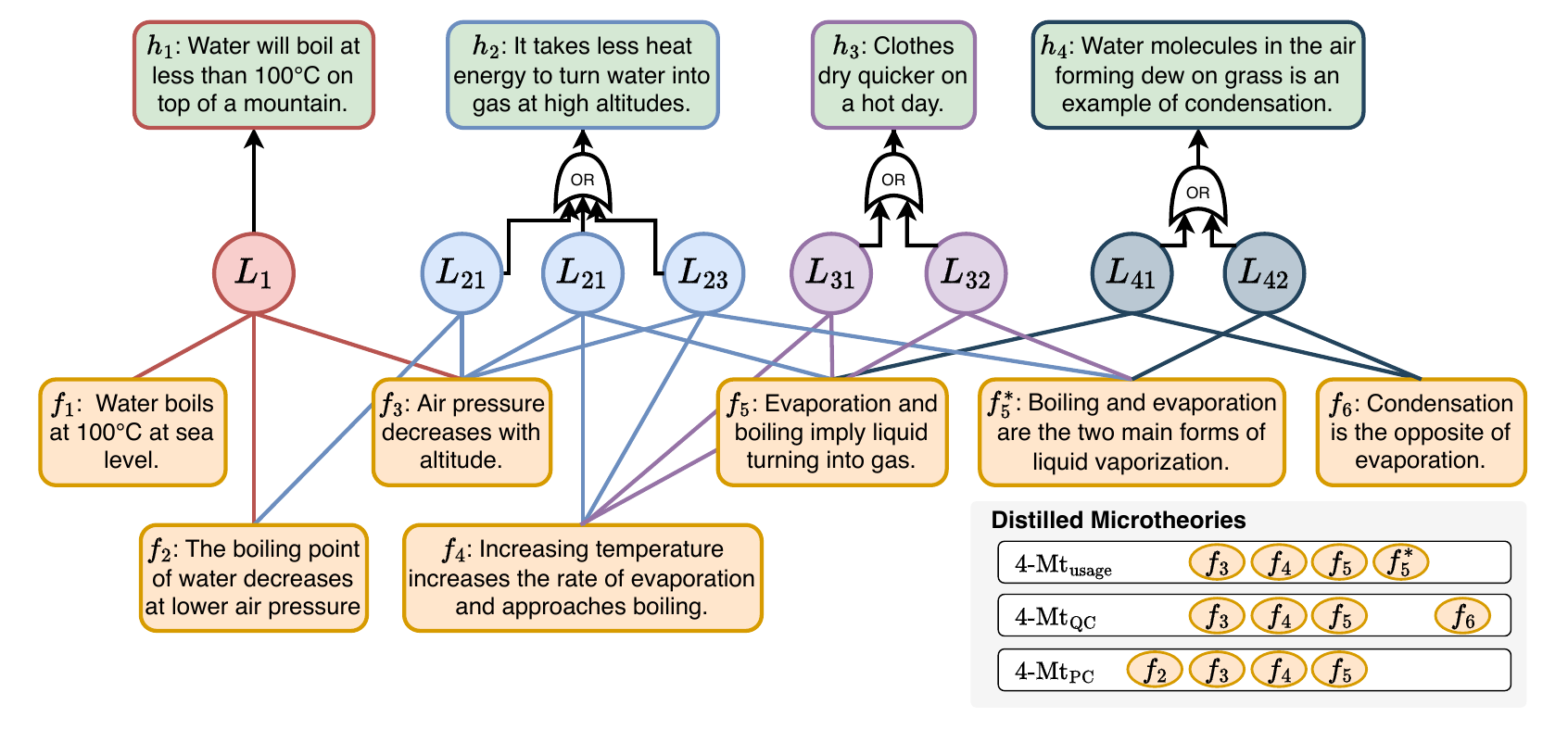}
    \vspace{-3mm}
    \caption{Comparison of distillation techniques for extracting $n$-Mts from a larger fact pool $\mathcal{C}$ based on their entailments ($L$s of different training hypotheses ($h$s)). The \textit{usage} approach prioritizes facts based on the number of hypotheses for which they are used. This risks keeping facts that serve the same role in explaining the same $h$s (e.g., $f_5$ and $f_5^*$ are both kept, even though only one is needed to explain $h_2$, $h_3$, and $h_4$). The \textit{question coverage} (QC) approach maximizes the number of $h$s for which the $n$-Mt contains {\it all} of a supporting argument's leaves (here $h_2$, $h_3$, and $h_4$). This risks failing to cover some $h$s at all (e.g. $h_1$). \textit{Partial coverage} (PC) maximizes the total {\it fraction} of the argument covered for all questions (preferring the most covered argument for each $h$, if there is more than one). In this example, coverage = .66+1+1+.5 = 3.16 (of a possible 4).
    }
    \label{fig:comp_approaches}
\end{figure}
\section{Experiments: Constructing Microtheories}
\label{sec:engine_exps}

We apply our method to construct microtheories for two domains: \textbf{ARC}~\cite{clark-etal-2018-think}, comprising grade-school science questions, and 
\textbf{MedQA}~\cite{jin-etal-2021-disease}, containing USMLE medical school exam questions.
Our goal is to extract generalizable knowledge statements using training questions that can be readily applied to solve test questions. This requires train/test splits that are about the same topics. 
To do this, we create mini-splits of each dataset containing topically similar questions with 6-900 training questions using a hierarchical topic clustering algorithm we describe in \S\ref{app:clustering}.
For ARC, we make a split with 9 topic clusters including ``Genetics and Evolution'', ``Force, Mass, Acceleration and Gravity'', and ``Plant Growth and Reproduction.''
For MedQA, we make a split with 4 clusters including ``The Effects of Smoking on Health'' and ``Kidney Function and Disorders.'' (See \autoref{tab:stats} for details).

\subsection{Microthery Construction Details}

To generate the initial fact pools (\S\ref{sec:extraction}),
we use GPT-4 (\texttt{gpt4-0613}). 
For entailment reasoning, we use \textsc{TreeWise}~\cite{weir-etal-2024-enhancing}, a recent, off-the-shelf, search-based SOTA 
entailment engine
that contains fine-tuned models to verify individual entailment steps (we use its tuned ChatGPT).
To demonstrate our work is not closed-model dependent, we show MedQA results using
Mixtral-8x22B-Instruct-v0.1~\cite{jiang-etal-2024-mixtral} in \sref{app:mixtral-results}.
\eat{
We use \textsc{TreeWise}~\cite{weir-etal-2024-enhancing} as our entailment engine, which is parametrized by an underlying LLM.
For ARC, We use GPT-4 (\texttt{gpt4-0613}) as the underlying LM to extract the Mts from training questions, and ChatGPT (\texttt{gpt-3.5-turbo-1106}) as the LM for test questions. We use the fine-tuned ChatGPT instance from \citet{weir-etal-2024-enhancing} to verify entailment steps.
For MedQA, we use Mixtral-8x22B-Instruct-v0.1~\cite{jiang-etal-2024-mixtral} as the underlying LM for both extraction and test-time inference.
}
We construct $n$-Mts for $n$=$25$ to $1000$ using the \textit{usage}, \textit{question coverage} and \textit{partial coverage} methods described in \S\ref{sec:methods}. 
Details are found in \sref{app:engine_config}.

\autoref{tab:stats} displays statistics about the number of questions and size of the resulting fact pools created during extraction. 
It shows that the redundancy reduction techniques in \sref{sec:entailment_condense} remove around 55\% of ARC facts and 25\% of MedQA.
\autoref{fig:fact_usage_hist} shows histograms for how frequently each fact in $\mathcal{C}$ was used in the 650-850 questions.
\begin{figure}[t!]
    \centering
    \begin{minipage}[t]{0.52\textwidth}
        \centering
        \adjustbox{valign=c} {\footnotesize
        \begin{tabular}{@{}lcc@{}}
        \toprule 
         & \textbf{ARC} & \textbf{MedQA} \\ \midrule
         \multicolumn{3}{@{}l}{\textbf{Dataset Details}} \\
        \# Questions     &  854/127/249  & 641/94/186 \\
        \# Topic Clusters     &  9 & 4 \\
        Ext. Corpus & Wikipedia & Wiki+Textbooks \\
        \midrule 
        \multicolumn{3}{@{}l}{\textbf{Training Extraction Results}} \\
        $|\mathcal{F}|$ & 20,561 & 16,722 \\
        $|\mathcal{C}|$ & 8,722 &  10,908  \\ 
        {\# Qs with Proof}          & 744 (89\%) & 562 (88\%) \\
        {Min \#Fs to Cover} & 800          & 890          \\
        {Fact/Q Ratio}              & 1.08         & 1.58        \\ \bottomrule
        \end{tabular}}
        \caption{(Upper) Dataset details for the two domains.
        (Lower) Results of microtheory extraction from training data. The last two rows result from the ``min-fact'' LP.
        }
        \label{tab:stats}
    \end{minipage}%
    \hfill
    \begin{minipage}[t]{0.45\textwidth}
        \centering
        \adjustbox{valign=c}{\includegraphics[trim={0 3mm 0 1mm},clip,width=\linewidth]{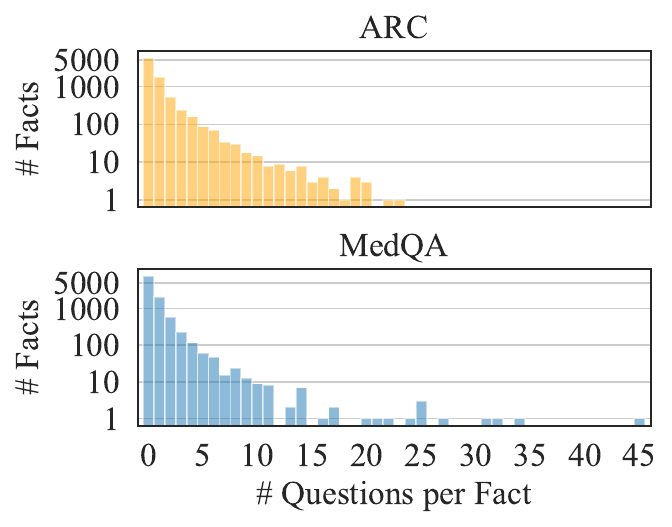}}
        \vspace{-2mm}
        \caption{Histogram of training question per fact in $\mathcal{C}$
        and 650-850 training questions. ARC facts are used more frequently than MedQA.}
        \label{fig:fact_usage_hist}
    \end{minipage}
\end{figure}
In both datasets, a substantial fraction of facts (5000/8000) are never used in a proof.
A few dozen facts from the ARC pool are used in more than 10 questions; fewer from the MedQA pool are used this frequently, suggesting that MedQA questions target more diverse pieces of knowledge than ARC, though a handful of MedQA facts are used in 30 or more questions.

To investigate this difference further
we implemented a ``min-fact'' linear program that found the smallest set of facts that fully covers a basis for every hypothesis the prover was able to ground. \autoref{tab:stats} shows that 
the 744 ARC training questions provable from $\mathcal{C}$
required a minimum of 800 facts for ``full coverage,'' a ratio of 1.08 facts per question, while MedQA 
required nearly twice that rate.
This highlights that the total amount of knowledge statements required to answer MedQA questions 
is substantially higher, which is unsurprising given the domains --
grade school- versus medical school-level topics.
We refer readers to \S\ref{app:qualitative} for  illustrations of the fact selection processes by $\text{Mt}_{\text{PC}}$ and $\text{Mt}_{\text{QC}}$,
examples of microtheory fact lists, and further qualitative analysis.

\subsection{Coverage of Training Hypotheses}
\label{sec:training_coverage}
How well do the usage-based optimization methods cover argumentative bases for the training hypotheses?
\autoref{fig:train_coverage} shows the fraction of full and partial coverage by each Mt approach.\footnote{We define fractional coverage of a question $Q_j$ as in \sref{sec:partial_coverage}: the highest fraction of leaves amongst all argumentative bases $L_{kj}$ found by the engine for the question.}
First, as expected, coverage increases with the size of the Mt.
Second, we see that our third optimization method, partial coverage (PC), performs best here,
with Mt$_\text{PC}$ (orange bars) consistently obtaining a higher coverage than either Mt$_\text{QC}$ or Mt$_\text{usage}$ (green, blue).
Finally, we observe full coverage is generally lower on MedQA than ARC, suggesting that
the MedQA domain contains fewer reusable principles and more idiosyncratic facts than ARC.
We later explore this conjecture further \S\ref{sec:relevance}.
\eat{
Coverage increases with the size of the Mt. Mt$_\text{PC}$ consistently obtains a higher coverage than either Mt$_\text{QC}$ or Mt$_\text{usage}$.
Mt$_\text{PC}$ finds a maximum coverage using 1000 facts in both domains\footnote{The entailment engine finds at least one basis for 88\% ARC and 74\% MedQA questions. } and achieves \~80\% of the maximum using half as many facts ($n=500$ vs $1000$).
Full coverage is overall substantially lower on MedQA than ARC, highlighting that 50 to 500 facts is not enough to fully explain the distribution of questions appearing in the 4 MedQA topics comprising the data.
}

\begin{figure}[t!]
    \centering
    \begin{minipage}[t]{0.47\textwidth}\vspace{0pt}
        \centering
        \begin{tabular}[t]{@{}c@{}}
            \includegraphics[trim={0 14.6mm 0 0},clip,width=\linewidth]{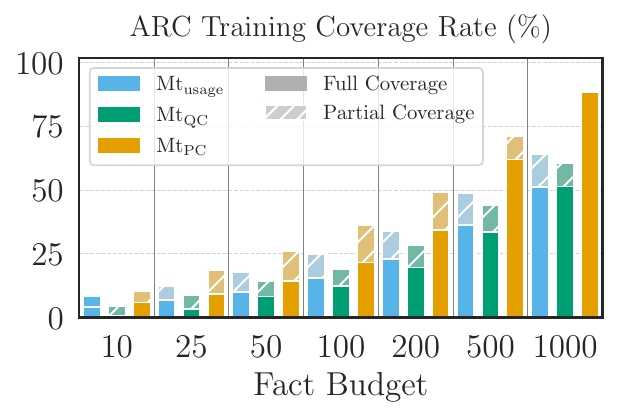} \\[\baselineskip]
            \includegraphics[trim={0 0 0 4pt},clip,width=\linewidth]{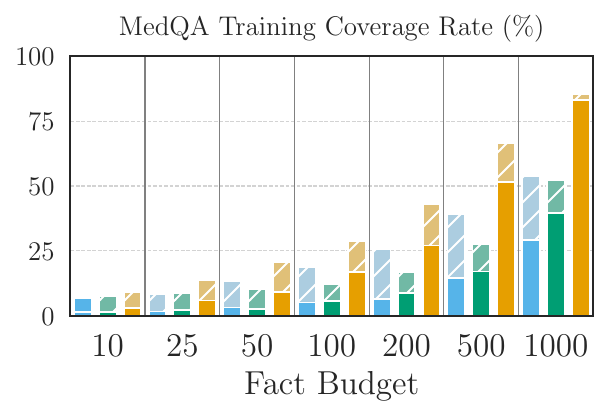}
        \end{tabular}
        \vspace{-5mm}
        \caption{Rate at which different microtheory approaches entail training hypotheses. 
        Lighter bars are the sum of fractional coverages for questions not fully covered by the Mt. Total coverage for ARC questions is generally higher than MedQA.
        The 1000-Mt$_{PC}$s effectively fully cover the training hypotheses for which a basis was found in both domains; since some hypotheses did not have a basis found, 100\% coverage is impossible.
        }
        \label{fig:train_coverage}
    \end{minipage}%
    \hfill
    \begin{minipage}[t]{0.51\textwidth}\vspace{0pt}
        \centering
        \begin{tabular}[t]{@{}c@{}}
            \includegraphics[trim={1mm 2mm 0 0},clip,width=\linewidth]{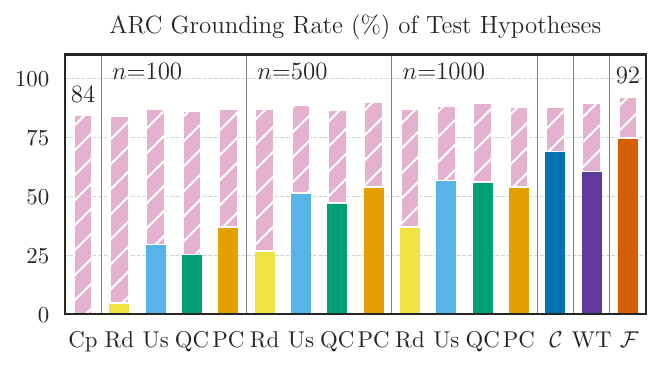} \\[\baselineskip] 
            \includegraphics[trim={0 0 0 4pt},clip,width=\linewidth]{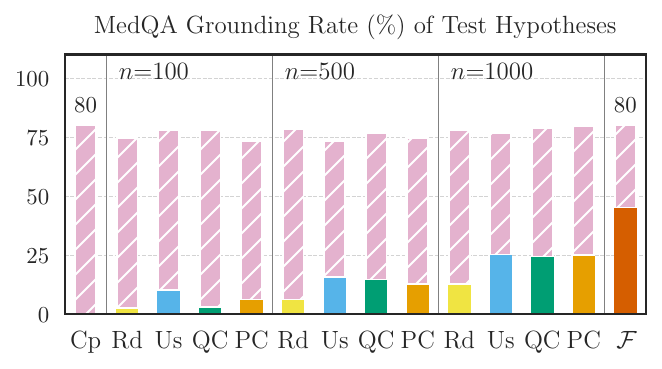}
        \end{tabular}
        \vspace{-3mm}
           \caption{
\% of test hypotheses fully groundable after adding various microtheories ($n$-sized \textbf{R}an\textbf{d}om, \textbf{Us}age, \textbf{QC}, \& \textbf{PC} Mts) to a base \textbf{C}or\textbf{p}us.
We show the fraction of leaves grounded in the Mt (solid) and the base corpus (striped). 
We see {\bf optimized Mts outperform the random baseline} (Rd), and trends show the tradeoff
of conciseness vs. performance. $\mathcal{F}$ {\bf provides up to +8\% grounding} over using just the corpus and exceeds the benefit of adding the handcrafted \textbf{W}orld\textbf{T}ree corpus.}
        \label{fig:test_coverage}
    \end{minipage}
\end{figure}

\section{Evaluation}
\label{sec:evaluation}

Our goal is that microtheories contain the core, reusable principles that underlie a topic.
Note that a microtheory (or any theory, for that matter) cannot contain {\it all} the knowledge
required to answer {\it all} topical questions, given that an infinite long-tail of
idiosyncratic facts is also required in addition to core principles. 
For example, to determine that a
dropped pencil (say) will fall to the ground requires not just general knowledge of gravity, but
also the idiosyncratic fact that a pencil is a denser-than-air object. To supply
such facts, we assess our microtheories when combined with a large, general corpus (Wikipedia and textbooks, see \S\ref{sec:test_grounding}),
treating each source as a flat retrieval index over knowledge statements. We evaluate along three dimensions:
\begin{enumerate}[topsep=0pt,itemsep=0ex,partopsep=1ex,parsep=1ex,labelindent=0pt,wide,labelwidth=!]
\item {\bf Grounding (\S\ref{sec:test_grounding}):} For the {\it correct} answers to the test questions, do microtheories
improve the number of answers that can be fully grounded in the (combined) corpus?
If so, this would suggest that
microtheories are supplying important, topical information not necessarily present in the
original corpus. We also measure what proportion of those groundings are supplied by
the microtheory vs. the original corpus, and the impact of Mt size.
\item {\bf Question Answering (\S\ref{sec:mt_qa_results}):} Using the entailment engine to answer the questions 
(i.e., selecting the entailed answer option, or the option with the highest-scoring entailment tree if more than one),
do microtheories improve QA accuracy? If so, this suggests Mts supply important
information to help a downstream task.
We evaluate this in our primary (science) domain, ARC.
\item {\bf Relevance (\S\ref{sec:relevance}): } To complement our empirical evaluations, we also perform a rigorous,
(human) expert evaluation of microtheory statements to assess, from a human perspective,
whether they appear to be critical, core domain principles (rather than idiosyncractic facts).
\end{enumerate}

\subsection{Grounding}
\label{sec:test_grounding}
We first evaluate, for the {\it correct} answers to test questions, whether
microtheories improve the number of answers that can be fully grounded by
the entailment engine (i.e.,
can be ``proved'' via entailment) when added to a base,
general corpus. For our primary domain (ARC), we use Wikipedia as our
base corpus, and for MedQA, we use a combination of Wikipedia and the set of medical textbooks released
by \citet{jin-etal-2021-disease}. These corpora are vastly
larger than the size of the microtheories.

\eat{
Here we use the entailment engine to evaluate whether Mts extracted for a set of training questions can ground unseen test hypotheses in the same topic.
We draw a distinction between \textit{covering} as in \sref{sec:training_coverage} and \textit{grounding} here:
unlike training questions, for which we prompted an LM to provide all the facts necessary to ground each hypothesis,
it is unlikely that an 1000-fact Mt alone will contain all idiosyncratic knowledge needed to ground a test hypothesis-- e.g. if the test hypothesis is about some drug not mentioned in training questions.
Therefore, instead of checking full coverage of an Mt alone, we test the effect of \textit{adding} Mts as additional grounding knowledge to a larger encyclopedic knowledge source. 
We use a retrieval index over Wikipedia (the same as used in \citet{weir-etal-2024-enhancing}) as our base corpus,\footnote{For MedQA, we include an additional base index over medical textbooks released by \citet{jin-etal-2021-disease}.} 
and then whether the added Mt, which is much smaller but contains specific and relevant knowledge, increases the rate of fully grounding test hypotheses to the combined corpora. 
}

\eat{
During entailment tree search, whenever \textsc{TreeWise} invokes its text retrieval module for a given hypothesis, we concatenate the results of retrieving from the Mt to the results from the external source(s), doubling the number of support documents at each search step.
}

\myparagraph{Baselines.}
We compare the $n$-Mt methods against a baseline that samples $n$ random facts from $\mathcal{F}$.
We also compare against using the entire pool $\mathcal{F}$.
For ARC, we also compare against the WorldTree resource, the version of which we use has 12,657 facts. We also consider using only the WorldTree facts that its annotators labeled as ``core'' for at least one question in their dataset (69\% of facts).

\myparagraph{Test Grounding Results.}
\autoref{fig:test_coverage} shows the hypothesis grounding rates when various microtheories are added to the
respective base corpora (Wikipedia, MedQA textbooks). 
A hypothesis is considered ``grounded'' if the engine found at least one entailment tree for it rooted fully in the Mt, corpus, and/or context.
Each bar is broken down into
the portions of trees grounded to the Mt (solid) vs. the base corpus (striped).\footnote{If the prover finds multiple trees, we take the one with the highest fraction of microtheory-rooted leaves.}
In both domains, comparing the first bar (Cp, base corpus only) and last bar ($\mathcal{F}$, using the full fact pool as a microtheory), we see grounding rates are substantially higher,
suggesting that {\bf microtheories can improve the the model's ability to ground its answers to verifiable knowledge}.
This capability is important, as it allows models to show how answers systematically follow from inspectable sources.

For ARC, the engine grounds up to 8\% more hypotheses (92\% vs. 84\%), and uses facts from $\mathcal{F}$ at a rate of 75\%
(solid part of the right-hand bar, i.e., 3/4 of the required entailment knowledge comes from the Mt).  In MedQA,
the hypotheses are grounded at a similar rate, but 45\% of the knowledge comes from the Mt instead of the corpus.
For ARC, we see a clear trend that as microtheory size (budget) increases,
the engine can ground more answers (overall bar heights), and ground to more of the Mt knowledge (solid bars).
For MedQA, the pattern is weaker: while larger Mts provide more grounding knowledge (solid bars),
there is no clear trend in overall grounding rates, suggesting that in this domain, there may be
fewer core, general principles vs. a large number of idiosyncratic facts.
We examine this
conjecture further in Section~\ref{sec:relevance}.

For ARC, we compare against a notable, manual attempt at a microtheory for ARC
called WorldTree, a corpus of several thousand statements (aiming to) cover all the knowledge
required for the ARC syllabus \cite{xie-etal-2020-worldtree}, constructed at significant expense
(\autoref{fig:test_coverage}, bar labelled WT). 
We see that \textbf{the n=1000 Mts result in similar rates of overall grounding
and grounding to the Mt yet were built automatically and are less than 10\% the size
of WorldTree.} 
Thus, our approach can be seen as automating this previously expensive process of
constructing such resources.

\subsection{Question Answering}
\label{sec:mt_qa_results}

\begin{wrapfigure}[18]{r}{0.54\textwidth}
    \centering
    \vspace{-10mm}
    \includegraphics[width=\linewidth]{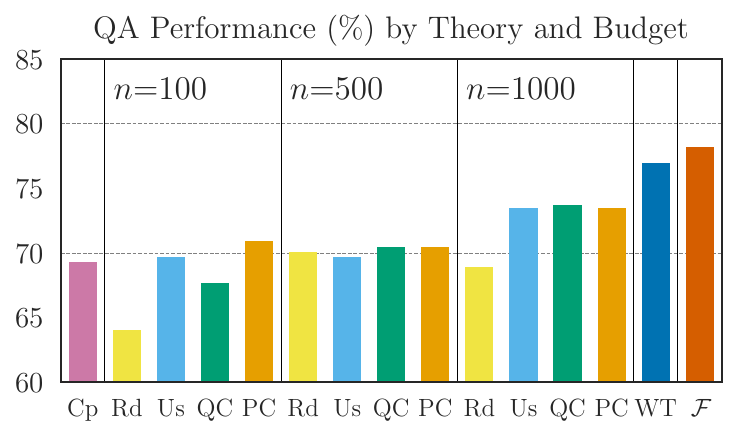}
    \vspace{-6mm}
    \caption{Question Answering performance on ARC topical test questions
    using Wikipedia plus various Mts (\textbf{R}an\textbf{d}om, \textbf{Us}age, \textbf{QC}, \textbf{PC}) as knowledge sources. $\mathcal{F}$ adds nearly 10\% over using only the base \textbf{C}or\textbf{p}us of Wikipedia despite being a tiny fraction of its size. This exceeds the benefits of the handwritten \textbf{W}orld\textbf{T}ree corpus.
    ($n$=1000)-Mts improve QA by 4\%. 
    This suggests that they contain important task information that is either not in Wikipedia or is hard to find.
    }
    \label{fig:mt_qa_results}
\end{wrapfigure}

\autoref{fig:mt_qa_results} shows the results of running multiple-choice question answering on 
the 9 ARC topics. For each question, the answer option that the entailment engine can ground in (``prove from'') the corpus is selected, or the option 
with the highest scoring entailment tree is selected if multiple options are entailed.
Using only Wikipedia as the knowledge source (Corpus, leftmost bar) achieves 69.2\% QA accuracy. Adding 100 or 500 facts with any approach generally does not improve on this and can decrease it using 100 random facts.\footnote{This decrease results from the irrelevant facts distracting the entailment search, hurting the grounding rate.}
When adding 1000-fact theories, all distillation methods add 4 points in QA (73\%). Adding the entire fact pool $\mathcal{F}$ further improves performance by 5 points (78\%).
These results suggest the following:

\begin{itemize}[topsep=0pt,itemsep=0ex,partopsep=1ex,parsep=1ex,labelindent=0pt,wide,labelwidth=!]
\item Adding Mts to Wikipedia shows a benefit to QA, despite their being orders of magnitude smaller.
    \item Mt size correlates with QA accuracy.
    For the ARC subset considered, one needs to retain at least 500 facts to see a noticeable improvement in QA accuracy.
    \item The question-driven Mt extraction method can provide performance gains on par with a clean, hand-annotated corpus (WorldTree) 
    \item When extracting core facts from $\mathcal{F}$, the choice of optimization method doesn't have a noticeable effect on QA, though each method outperforms random selection.
\end{itemize}

\subsection{Relevance}
\label{sec:relevance}

Finally, to complement our empirical investigation, we perform two
subjective assessments of the contents of the microtheories, the first using
(expert) human experts and the second using an LLM-as-judge. This provides an 
orthogonal evaluation of whether our microtheory-building process is indeed
extracting the more important, central facts about the topic. We are 
particularly interested in whether, and by how much, our microtheory condensation and
optimization methods (\S\ref{sec:usage_optimization}) are selecting important facts from the 
general fact pool $\mathcal{F}$.

\eat{
In this project, we have proposed a \textit{question-driven} approach to constructing a microtheory,
relying on the assumption that
the knowledge needed to answer a training set
is the same needed to answer test questions.
In this section, we explore whether this assumption holds for our domains, or whether the discrepancy in required knowledge helps explain the 
findings in \sref{sec:engine_exps} that Mts improve hypothesis grounding rates and QA in the ARC domain, but not always in the MedQA domain.
In \sref{sec:human_relevance} we collect expert judgments to measure whether Mts contain facts that are generally relevant for humans to know for MedQA exams.
In \sref{sec:auto-relevance} we introduce an automatic scoring mechanism for whether an Mt is relevant on a per-question basis. We use the mechanism to then introduce a dataset-level metric in \sref{sec:par-metric} quantifying the extent to which an Mt for any training dataset will be relevant to test questions in the same distribution.
}

\subsubsection{Expert Relevance Annotations}
\label{sec:human_relevance}
First, we collect human judgments of the relevance of Mt facts.
We use the MedQA medical domain due to easy access to domain experts.
We asked experts how useful they would find the Mt facts when studying for and taking USMLE tests. 
We recruited two senior medical students to annotate the general relevance of facts from different MedQA microtheories.
\begin{wrapfigure}[16]{r}{0.49\textwidth}
    \centering
    \vspace{-4.5mm}
    \includegraphics[width=\linewidth]{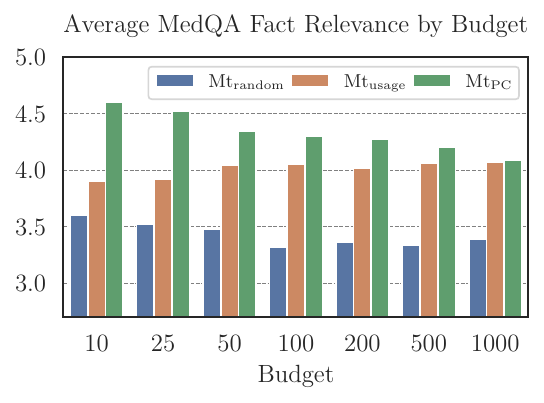}
    \vspace{-8mm}
    \caption{Average expert-annotated task relevance score for microtheory facts.
    \textit{PC} facts are more relevant than \textit{random} and \textit{usage} until the 1000-fact budget.
    }
    \label{fig:relevance_expert_results}
\end{wrapfigure}
We worked with the annotators to develop a 5-point rubric for scoring how essential a fact is 
to the core content that appears on the USMLE.
Annotation details and the full rubric are in \sref{app:expert-ratings}.

Average relevance ratings are shown in \autoref{fig:relevance_expert_results}.
Both optimization approaches outperform a random baseline by at least half a grade point; Mt$_{\text{PC}}$ is 0.6-1.2 higher,
and never drops below a score of 4.
(The annotation rubric score defines a score of 4 as containing relevant exam content, though either too general or unspecific to sufficiently answer an exam question.) In contrast, a randomly selected fact from the full pool frequently scores below 3.5, often containing context or background information less directly useful for answering exam questions.
This suggests that the Mt extraction methods successfully
distill out the most topically relevant facts from the general knowledge pool, successfully selecting the information most likely to support exam performance.

\subsubsection{Automatic Per-Question Relevance}
\label{sec:auto-relevance}

As well as evaluating the Mt as a whole, we also explore whether Mts are relevant on a {\it per-question} basis,
to see how often the Mt has {\it something} useful to contribute when answering a new question.
We ran a study to assess this, using an LLM as a judge. Although
necessarily approximate using an LLM as the scorer, it provides further insights into our Mts.
First, we say a Mt is {\it relevant} if it contains
{\it at least one} fact that is relevant for getting the question right, i.e., differentiating
the correct answer from any incorrect options. To determine if a fact is relevant,
we use an existing scoring rubric developed by \citet{jansen-etal-2021-challenges} for rating a fact's relevance to a question
on a 0-5 scale, and here ask \texttt{gpt-4o-2024-08-06} to make that judgment (see \S\ref{app:auto_relevance_prompt}
for the rubric and the full prompt). We then measure the rate at which at least one 5-rated fact is
in the Mt.\footnote{For practical reasons, we only assess the top 270 facts retrieved from the Mt by the entailment engine (for Mts larger than 270). The engine uses an SBERT retrieval encoder fine-tuned on science QA support facts. }
This provides
a soft indicator that the Mt contains information that would be core
to a basis for the correct answer. %

\eat{
The previous section found that Mts are \textit{generally} relevant for a medical student to know for USMLE exams; here we explore whether they are relevant on a \textit{question-to-question} basis.
We construct an LLM-based metric that lower-bounds a Mt's relevance to a given question by checking whether it contains \textit{at least one} fact that is directly relevant to getting the question right, i.e. differentiating the correct answer from any incorrect options.
We query the \treewise{} support fact retrieval index for each question to obtain some top $k$=$270$ facts relevant to the correct answer.
We then prompted GPT-4\footnote{\texttt{gpt-4o-2024-08-06}} to rate each retrieved fact on an ordinal relevance scale based on the rubric used by \citet{jansen-etal-2021-challenges} for expert annotators to score the relevance of WorldTree facts to questions.
The prompt used for scoring facts is shown in \S\ref{app:relevance_prompt}. 
We measured the rate at which at least one 5-rated fact appeared in the retrieved fact set, using this as a soft indicator that the Mt contains \textit{any} useful information that would be core to an argumentative basis for the correct hypothesis option. 
}

\autoref{fig:auto_relevance_results} shows the change in per-question relevance rate as we increase $n$ for Mt$_{\text{PC}}$. 
Our main observations are that \textbf{Mt$_{\text{PC}}$ is substantially more relevant than Mt$_{\text{random}}$} (see \autoref{fig:relevance_results} for other methods) and that the \textbf{ARC Mts contain 5/5 relevant facts at a much higher rate than MedQA.}
For example, the 100-Mt$_{\text{PC}}$ obtains a 72.3\% relevance rate for ARC, while for MedQA,
this rate drops to 30.1\%. Similarly, the 1000-Mt$_{\text{PC}}$ has around 90\% relevance for ARC but
only 70\% for MedQA. Even looking at the entire fact pool $F$, it is only 93\% relevant
for MedQA (right-hand bar in Figure~\ref{fig:auto_relevance_results}), while nearly 100\% relevant for ARC.
This suggests that it is easier to find highly reusable, general principles in ARC's domain
(science), e.g., basic laws of electricity, than in MedQA (medicine). Discussions
with our medical experts reinforce this: while there are also general principles in
medicine, e.g., ``The body strives to maintain a stable internal environment,''
these are fewer and far between, and the USMLE questions instead often probe a
student's long-tail knowledge of highly specific medical facts (e.g., symptoms of
Hippel-Lindau disease). This suggests that some domains may be more amenable to distilling into
microthoeries than others, and explains the more limited gains seen
earlier with MedQA compared with ARC.

\eat{ %
The ARC Mts contain 5/5-relevant facts at a much higher rate than MedQA. 50-Mt$_{\text{PC}}$ obtains a 56\% relevance rate for ARC while 100-Mt$_{\text{PC}}$ obtains 72.3\%. For MedQA, these rates decrease to 20.4\% and 28.5\%. 1000-Mt$_{\text{PC}}$ have around 90\% relevance for ARC but only 55-67\% for MedQA. The fact pool $\mathcal{F}$ and condensed pool $\mathcal{C}$ are nearly 100\% relevant for ARC but only 85\% for MedQA.}
\eat{
This provides a possible explanation for the ineffectiveness of the MedQA Mts: they are \textit{precise} (generally relevant) but have low \textit{recall} (per-question irrelevant).
At best, the largest 1000-fact-or-less Mts contain at least one relevant fact only 2/3 of the time. If the entailment engine is provided with Mt facts during decompositional search, they are likely not helpful in discerning correct answer hypotheses.
}

\begin{figure}[t!]
    \centering
    \begin{minipage}[t]{0.49\textwidth}\vspace{0pt}
        \centering
        \includegraphics[trim={1mm 0 1mm 0},clip,width=\linewidth]{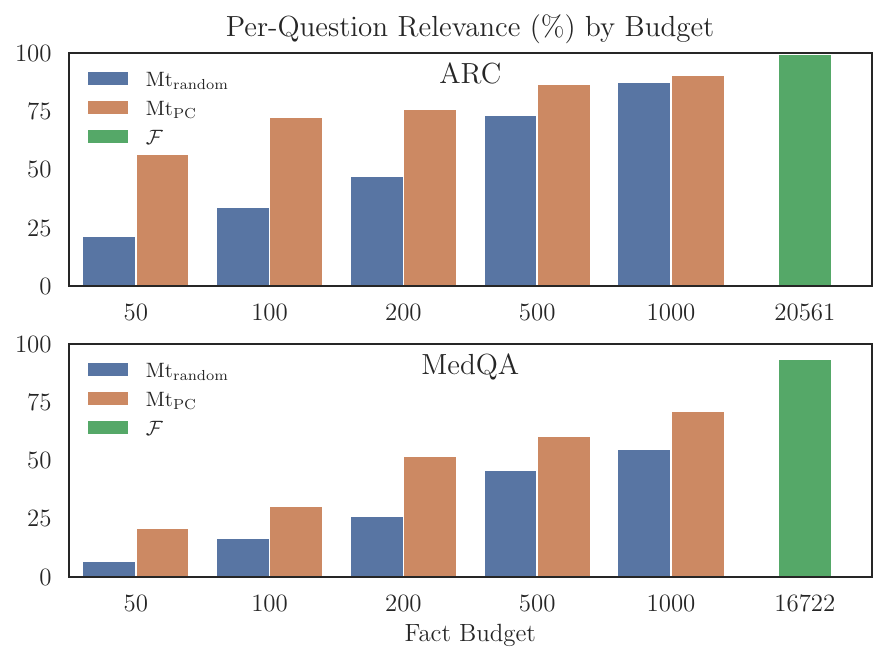}
        \vspace{-7.5mm}
        \caption{
        Per-question relevance rate for $n$-MT$_\text{PC}$s for varying $n$s.
        The rate is much higher in the ARC domain than MedQA; at 1000 facts, the Mt is \textit{at all} relevant to only 2/3 of MedQA questions but 90\% of ARC. 
        }
        \label{fig:auto_relevance_results}
    \end{minipage}%
    \hfill
    \begin{minipage}[t]{0.49\textwidth}\vspace{0pt}
        \centering
        \includegraphics[trim={4mm 0 4mm 0},clip, width=\linewidth]{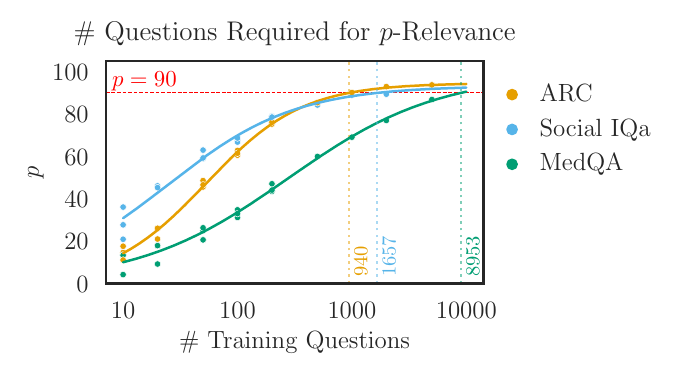}
        \vspace{-8mm}
        \caption{Number of training questions (x-axis) needed to create an $\mathcal{F}$ that is relevant to test questions $p$\% of the time (y-axis).
        While datasets like ARC and Social IQa~\cite{sap-etal-2019-social} need 900-1600 questions for 90-relevance,
        but MedQA needs far more (8900). 
        }
        \label{fig:p_relevance}
    \end{minipage}
\end{figure}

\section{How Complete are Microtheories?}
\label{sec:par-metric}

How many training questions do we need to construct a ``complete'' microtheory?
This is an important, practical concern for anyone wanting to extract microtheories
for a new domain. In one sense, a microtheory is never complete,
as there will always be topical questions that touch areas uncovered in training. However, 
we can make progress on this question in a different way. First, 
given 
a distribution over topical questions,
we say that a microtheory is {\it relevant} to a question
if it contains at least one fact
that is core to the (correct) answer
according to the method described in \sref{sec:auto-relevance}.
We then say a microtheory is partially ($p$ = e.g., 90\%) complete
if it is relevant to an arbitrary sampled question with probability $p$. Finally, we
can predict this {\it $p$-relevance} by generating and extrapolating from microtheory
learning curves such as in \autoref{fig:p_relevance}.\footnote{We fit ($ R^2 = 1.00$) a modified Hill equation with an additional scaling parameter:
$y = \frac{V_{\text{max}} \cdot x^n}{K^n + x^n} \cdot s$. }
For example, to obtain an Mt
with a $p$-relevance of 90\%, we will likely need 940 training questions for ARC,
but 8953 training questions for MedQA.\footnote{Unlike in \autoref{fig:auto_relevance_results}, we use the original train/test sets here. 
We compute the per-question relevance of $\mathcal{F}$ prior to any reductions, thus posing an upper bound on the relevance of any Mt distilled from the larger pool.}
Conversely, given a training set of size $n$,
we can predict the likely $p$-relevance of the Mt from the curve.
Thus we would expect 100 questions to provide enough knowledge to be ($p$=60)-relevant for ARC but only 30-relevant for MedQA.

$p$-relevance gives practitioners a way of predicting whether Mts will be helpful for their new dataset:
determining how much data they would need to collect for a new domain for the Mt to be frequently useful for the topic. If this number is too high, then Mts might not be a viable solution for the domain -- this is likely the case for many datasets for which there is less of a shared body of knowledge that is reapplied to many questions, e.g. HotPotQA~\cite{yang-etal-2018-hotpotqa} and Natural Questions~\cite{kwiatkowski-etal-2019-natural}, where each question asks about a unique Wikipedia factoid. 
The number of training questions also has a large effect on the amount of compute used to construct a microtheory, since calling the entailment engine to collect usage statistics and distill the most generalizable facts requires many LLM calls per question. 
The smaller the training set necessary to obtain high $p$ values, the less LLM calls;
running the engine on 8953 MedQA questions to obtain 90-relevance would be making 100,000s of LLM calls. 
Future work might explore which kinds of QA datasets are realistically conducive to Mts.

\section{Conclusion}

While LLMs have strong reasoning abilities, it is often unclear how their beliefs recurringly
interact to answer questions on a topic, a significant barrier for understanding and deploying models.
To address this, we have presented a method for materializing a model's latent,
topical knowledge into a {\it microtheory}, a linguistic analog of logical microtheories,
articulating the model's core, reusable knowledge about the topic. In contrast to
explanation methods that show snippets of knowledge supporting individual answers,
ours is the first method that attempts to distill a model's {\it overall understanding}
of a topic. Our evaluations suggest that, when added to a general corpus, microtheories
can significantly improve a model's ability to ground its answers, i.e.,
show how they are systematically entailed by corpus documents, as well
as improving the accuracy of those grounded answers. In addition, human and LLM-as-judge
assessments of the microtheories suggest they contain a high proportion of
critical, topical facts, and that our construction process successfully
identifies many of those important facts from a general pool. Together,
these suggest that microtheries are an efficient distillation of model's
topic-relevant knowledge, and can provide both performance
gains and, for the first time, an interpretable window into a model's topical knowledge.

\eat{
While LLMs have strong reasoning abilities, 
we are not sure how their beliefs
recurringly interact in order to answer multiple different questions on the topic. 
In this paper, we explored the automatic construction of \textit{microtheories}: concise and interpretable lists of core knowledge statements that reflect an LM's internal beliefs and are readily applied to support answers to questions in a dataset.
We provide a series of methods for extracting knowledge from LLMs and distilling it down to the core principles using recent advances in neural textual entailment reasoning.
We explore the benefits of our new explicit belief representation for grounding new domain hypotheses via textual entailment, showing that it can improve the performance of entailment-based reasoning—matching or exceeding the benefits of similar, hand-crafted knowledge resources. 
We also perform a series of human and automatic evaluations that provide evidence that microtheories frequently contain the necessary relevant information for solving test questions, and introduced a framework for analysis, called $p$-relevance, for practitioners to use when determining how microtheories might benefit their new problem domains. 
}

\bibliographystyle{iclr2025_conference}
\bibliography{anthology-shrunk-1-24, aaai24}
\pagebreak
\appendix

\section{Usage Optimization Details}
\subsection{$n\text{-Mt}_\text{QC}$ Linear Program}
\label{app:qc_ilp}
We set the objective function to achieve the following goals, ordered by precedence:
\begin{itemize}
    \item Maximize the coverage of questions by at least one proof tree.
    \item Maximize the number of proof trees covered by the selected facts.
    \item Minimize the number of facts included in the new theory.
\end{itemize}
We define the objective as 
\begin{equation}
\min \left( \sum_{i=1}^{|\mathcal{C}|} x_i - \sum_{j=1}^{d} \left( \omega z_i  + \sum_{k=1}^{|\mathcal{L}_i|} y_{kj} \right)   \right)
\end{equation}
Where $x_i$ is a binary variable indicating whether fact $i$ is included in the Mt, $y_{kj}$ is a binary variable indicating whether basis $L_{kj}$ for question $j$ is covered by the selected facts, and $z_j$ is a binary variable indicator whether question $q_j$ has at least one basis fully covered. We set $\omega$ to be sufficiently high to dwarf the other terms, using them only for breaking ties.

We impose the following constraints:
\begin{itemize}
    \item At most $n$ facts are kept.
\begin{equation}
\label{eqn:budget}
    \sum_{i=1}^{|\mathcal{C}|} x_i \leq n
\end{equation}
\item Each $L_{kj}$ is ``covered'' iff all its facts are kept.
\begin{equation}
    \forall L_{kj}, \quad \sum_{i = 1}^{|L_{kj}|} x_i \geq y_j \cdot |L_{kj}|
    \end{equation}
\item Each question is ``covered'' if at least one of its bases is ``covered''
\begin{equation}
    \forall \mathcal{L}_j, \quad \sum_{j = 1}^{| \mathcal{L}_j|} y_j \geq z_k
\end{equation}
\end{itemize}

\subsection{$n\text{-Mt}_\text{PC}$ ILP Implementation}
\label{app:pc_ilp}
We set the objective function to achieve the following goals, ordered by precedence:
\begin{itemize}
    \item Maximize the partial coverage of each question by the selected facts.
    \item Minimize the number of facts included in the new theory.
\end{itemize}
We define the objective as 

\begin{equation}
\min \left( \sum_{i=1}^{|\mathcal{C}|} x_i - \sum_{j=1}^{d} \omega p_j   \right)
\end{equation}

Where $x_i$ is a binary variable indicating whether fact $i$ is included in the Mt, $p_j$ is a continuous variable indicating the maximal partial coverage of question $j$ by the selected facts, and $y_{kj}$ is a continuous variable indicating the coverage of basis $L_{kj}$ for question $j$ by the selected facts. $z_{kj}$ is a binary indicator of whether a given $L_{kj}$ is the highest coverage of question $j$ attained by the solution.

We impose the following constraints:
\begin{itemize}
    \item At most $n$ facts are kept using \autoref{eqn:budget}.
\item Each $L_{kj}$ is ``covered'' by the proportion of its facts that are kept.
\begin{equation}
    \forall L_{kj}, \quad y_{kj} = \frac{1}{|L_{kj}|} \sum_{i \in L_{kj}} x_i
\end{equation}
\item Two constraints that enforce $p_j$ equals the highest parital coverage for question $j$ using the Big M method: The maximal partial coverage of each question is at least as large as the coverage of each of its bases.
\begin{equation}
\label{eqn:partial_coverage_one}
    \forall \mathcal{L}_j, \quad p_j \geq y_{kj} - M(1 - z_{kj}) \quad \forall k
\end{equation}
and the maximal partial coverage of each question does not exceed the coverage of each of its bases.
\begin{equation}
\label{eqn:partial_coverage_two}
    \forall \mathcal{L}_j, \quad p_j \leq y_{kj} + Mz_{kj} \quad \forall k
\end{equation}
\item Exactly one $z_{kj}$ is active per question.
\begin{equation}
    \forall \mathcal{L}_j, \quad \sum_{k = 1}^{| \mathcal{L}_j|} z_{kj} = 1
\end{equation}
\end{itemize}

\section{Question/Topic Clustering Algorithm}
\label{app:clustering}
To create our topic-specific mini-splits of ARC and MedQA, we use a hierarchical clustering algorithm that gathers questions targeting similar knowledge by iteratively prompting an LLM to `caption' questions with core targeted knowledge sentences and then clustering them using SBERT.
Algorithm \ref{alg:two_step_clustering} depicts our implementation of the question clustering algorithm used to extract subsets of a broader QA dataset that target similar underlying core knowledge. 
For the embedding clustering algorithm we use \href{https://github.com/UKPLab/sentence-transformers/blob/179b659621c680371394d507683b25ba7faa0dd8/sentence_transformers/util.py#L346}{the SentenceTransformer-based community detection algorithm} that clusters sentences greedily according to some minimum cosine similarity threshold, which we set to $0.6$. 

\begin{algorithm}[ht!]
    \begin{algorithmic}[1]
        \Procedure{\textcolor{blue}{TwoStepClustering}}{$Q$}
            \ra{} Hierarchically clustered questions with topic and hypertopic labels:
            \State $core\_facts$ = []
            \State \textbf{forall} $q_i \in Q$ \textbf{do}
            \State \gapxx $core\_fact = \textcolor{blue}{\textproc{LLM.ExtractCoreFact}}(q_i)$
            \State \gapxx $core\_facts$ += $core\_fact$
            \State \textbf{end for}

            \State $clusters = \textcolor{blue}{\textproc{SBERT.Cluster}}(core\_facts)$
            \State $topic\_labels$ = []
            \State \textbf{forall} $c_j \in clusters$ \textbf{do}
            \State \gapxx $topic\_label = \textcolor{blue}{\textproc{LLM.GenerateTopicLabel}}(c_j)$
            \State \gapxx $topic\_labels$ += $topic\_label$
            \State \textbf{end for}

            \State $topic\_clusters = \textcolor{blue}{\textproc{SBERT.Cluster}}(topic\_labels)$
            \State $hypertopic\_labels$ = []
            \State \textbf{forall} $tc_k \in topic\_clusters$ \textbf{do}
            \State \gapxx $hypertopic\_label = \textcolor{blue}{\textproc{LLM.GenerateHypertopicLabel}}(tc_k)$
            \State \gapxx $hypertopic\_labels$ += $hypertopic\_label$
            \State \textbf{end for}

            \State \textbf{Assign $topic\_labels$ and $hypertopic\_labels$ to corresponding questions in $Q$}

            \State \textbf{return} $Q$ with hierarchical topic and hypertopic labels
        \EndProcedure
    \end{algorithmic}
    \caption{Two-Step Clustering Algorithm for Question Grouping by Core Topic}
    \label{alg:two_step_clustering}
\end{algorithm}

\section{Fact Extraction Prompting Details}
\label{app:theorycot}
The LLM prompt to generate relevant facts for a given $Q_i$ is a few-shot learning prompt using exemplars dynamically retrieved\footnote{We use BM25~\cite{robertson-etal-1995-okapi} for retrieval using the question text as the query.} from EntailmentBank. Each exemplar's fact list comprises the gold WorldTree~\cite{xie-etal-2020-worldtree} facts from the unstructured explanation graph, and its entailment tree is the gold entailment tree that uses a subset of the facts.
Thus, while $Q_i$ might not be in the same ScienceQA domain as the exemplars, we cue the LM to generate \textit{WorldTree-like} facts that should serve similar roles as argumentative bases in entailment trees.
We discard the model-generated entailment tree and take the generated fact list as candidates to be added to the fact pool. 

\autoref{fig:theorycot-icl} shows an example dynamically constructed in-context learning prompt for fact extraction. 
\autoref{fig:theorycot-outputs} shows an example output from Mixtral-8x22b-Instruct-V0.1 for a MedQA question.

\begin{figure}[H]
    \centering
    \scriptsize
    \begin{lstlisting}[style=base,basicstyle=\tiny\ttfamily]
You are an expert theory generation system that lists facts about a problem area that will help solve a reasoning question. Given a question, you generate QUERY statements for each of the possible answers to the question. The queries should be complete statements that exactly match the answer choices as closely as possible.

Then, you generate a theory comprising simple FACTS about the world. These should be statements that you believe to be logical and true about the world that will help you prove the queries. A FACT is either a generic statement about the world ("birds can fly") or a specific statement about the context that helps support the proof of the query ("Tweety is a bird").

For a QUERY, you generate at least 6 statements.

Once you generate the theory, you show how one of the QUERYs logically follows from the context. The selected QUERY corresponds to the answer choice that you believe is correct. You can only choose one QUERY as correct.

To show your reasoning, you generate a PROOF of your chosen QUERY hypothesis. Your PROOF is an "entailment tree" that composes the FACTS of your theory in order to show how the query is compositionally entailed by them. Each step in the tree is a composition of two or more premises into a new conclusion, which should be used as a premise in future steps. The last step compositionally proves the hypothesis. When you reuse the new conclusion in a later step, you refer to it by its id. Each step is separated by a newline.

###
QUESTION: Which of the following actions is most likely part of a test to find the hardness of a mineral sample? (A) heating the sample on a hot plate (B) scratching the sample with a nail (C) hitting the sample with a hammer (D) shining a bright light on the sample

QUERIES:
(QUERY A) Heating the sample on a hot plate is most likely part of a test to find the hardness of a mineral sample.
(QUERY B) Scratching the sample with a nail is most likely part of a test to find the hardness of a mineral sample.
(QUERY C) Hitting the sample with a hammer is most likely part of a test to find the hardness of a mineral sample.
(QUERY D) Shining a bright light on the sample is most likely part of a test to find the hardness of a mineral sample.

THEORY:
(FACT 1) a mineral is a kind of material
(FACT 2) hardness is a measure of a mineral 's ability to resist scratching
(FACT 3) an example of breaking something down is scratching something
(FACT 4) a material that is soft can be broken down by a material that is hard
(FACT 5) testing often requires measuring
(FACT 6) a nail is a kind of object
(FACT 7) hardness is a property of a material / an object and includes ordered values of malleable / rigid
(FACT 8) hardness can be used to identify minerals
(FACT 9) a mineral is a kind of solid / natural material
(FACT 10) nails are often made of iron
(FACT 11) scraping an object may cause small particles to break off of that object
(FACT 12) if a mineral can be scratched by a fingernail then that mineral is soft
(FACT 13) hardness is a kind of physical property
(FACT 14) hardness is an intensive property
(FACT 15) soft is a kind of hardness
(FACT 16) experimentation requires measuring
(FACT 17) an event is a kind of action
(FACT 18) if a material scratches easily then that material has low hardness
(FACT 19) a mineral is a kind of object
(FACT 20) mineral nutrient is a kind of nutrient
(FACT 21) objects are made of materials / substances / matter
(FACT 22) rock is made of minerals
(FACT 23) a process is made of a series of actions
(FACT 24) if an object is made of a material then that object has the properties of that material
(FACT 25) testing / identifying the streak of a mineral requires the powdered form of that mineral


PROOF of Scratching the sample with a nail is most likely part of a test to find the hardness of a mineral sample.:
(STEP 1) a material that is soft can be broken down by a material that is hard & an example of breaking something down is scratching something -> int1: a material that is soft can be scratched by a material that is hard
(STEP 2) int1 & a mineral is a kind of material -> int2: if one mineral can scratch another mineral then that other mineral is softer than that one mineral
(STEP 3) hardness is a measure of a mineral 's ability to resist scratching & int2 -> hypothesis: Scratching the sample with a nail is most likely part of a test to find the hardness of a mineral sample.

ANSWER: B

### 
QUESTION: {question}

QUERIES:
{queries}

THEORY:
    \end{lstlisting}
    \caption{TheoryCoT in-context learning prompt for extracting WorldTree-like facts from an LLM for a given question.}
    \label{fig:theorycot-icl}

\end{figure}

\begin{figure}[t!]
    \centering
    \scriptsize
    \begin{lstlisting}[style=base,basicstyle=\tiny\ttfamily]
### 
QUESTION: A 47-year-old woman with chronic epigastric pain comes to the physician because of a 1-month history of intermittent, loose, foul-smelling stools. She has also had a 6-kg (13-lb) weight loss. She has consumed 9-10 alcoholic beverages daily for the past 25 years. Seven years ago, she traveled to Mexico on vacation; she has not been outside the large metropolitan area in which she resides since then. She appears malnourished. The stool is pale and loose; fecal fat content is elevated. An immunoglobulin A serum anti-tissue transglutaminase antibody assay is negative. Further evaluation is most likely to show which of the following? (A) Trophozoites on stool microscopy, (B) Pancreatic calcifications, (C) Villous atrophy of duodenal mucosa, (D) Positive lactulose breath test

QUERIES:
(QUERY A) Further evaluation is most likely to show trophozoites on stool microscopy.
(QUERY B) Further evaluation is most likely to show pancreatic calcifications.
(QUERY C) Further evaluation is most likely to show villous atrophy of duodenal mucosa.
(QUERY D) Further evaluation is most likely to show a positive lactulose breath test.

THEORY:
@(FACT 1) Chronic epigastric pain can be a symptom of various gastrointestinal disorders.
(FACT 2) Intermittent, loose, foul-smelling stools can be a sign of malabsorption.
(FACT 3) Significant weight loss can be a sign of malnutrition or malabsorption.
(FACT 4) Long-term, heavy alcohol consumption can lead to pancreatitis and malabsorption.
(FACT 5) Travel to certain regions can expose individuals to parasites that cause gastrointestinal issues.
(FACT 6) A negative immunoglobulin A serum anti-tissue transglutaminase antibody assay rules out celiac disease.
(FACT 7) Malabsorption can lead to elevated fecal fat content.
(FACT 8) Pale and loose stools can be a sign of malabsorption.
(FACT 9) Pancreatic calcifications can be a sign of chronic pancreatitis.
(FACT 10) Villous atrophy of duodenal mucosa can be a sign of celiac disease or other gastrointestinal disorders.
(FACT 11) A positive lactulose breath test can indicate small intestinal bacterial overgrowth (SIBO).

PROOF of Further evaluation is most likely to show pancreatic calcifications.:
(STEP 1) Long-term, heavy alcohol consumption can lead to pancreatitis and malabsorption & Significant weight loss can be a sign of malnutrition or malabsorption -> int1: The patient's long-term, heavy alcohol consumption may have led to pancreatitis and malabsorption, causing significant weight loss.
(STEP 2) int1 & Intermittent, loose, foul-smelling stools can be a sign of malabsorption -> int2: The patient's intermittent, loose, foul-smelling stools are likely due to malabsorption caused by pancreatitis.
(STEP 3) int2 & Pancreatic calcifications can be a sign of chronic pancreatitis -> hypothesis: Further evaluation is most likely to show pancreatic calcifications.@
    \end{lstlisting}
    \caption[Example TheoryCoT output for a MedQA question.]{Example TheoryCoT output for a MedQA question. The generated proof steps are thrown out and the generic facts are added to the fact pool $\mathcal{F}$. Context-specific facts are discarded. }
    \label{fig:theorycot-outputs}
\end{figure}

\section{Entailment Engine Implementation Details}
\label{app:engine_config}
To collect training proofs, we explore up to depth 2 for ARC hypotheses (these are easier to prove) and up to 40 expansions for MedQA. 
\subsection{Retaining Inferences}
\label{sec:retaining}
When collecting the bases $L_{ij}$, we want the entailment engine to best identify which facts are most commonly reused across questions. 
Since the engine has the potential to miss certain combinations, we add a mechanism to encourage reusing inferences from earlier training questions $Q_{1}\dots Q_{i-1}$ when finding bases for $Q_{i}$. 
After each call to $\textsc{engine}(h_i, \mathcal{C}; Q_i))$, we take all intermediate sub-hypotheses grounded by the engine to facts in $\mathcal{C}$ in the process of proving $h_i$, some $\mathcal{S}_{i}$, and add them to the fact pool for all future questions $j > i$.
Whenever the engine uses any subhypothesis $s \in S_i$, we replace it with all the ways the engine grounded $s$ using bases in $\mathcal{C}$. E.g., if the engine returns basis $[f_1, s]$ and $s$ was itself previously grounded by the sets $[f_2, f_3], [f_2, f_4]$, we replace $[f_1, s]$ with $[f_1, f_2, f_3]$  and $[f_1, f_2, f_4]$.\footnote{To limit combinatorial explosions, we produce a maximum of 1000 bases per question via this ``unfolding''.}

\section{Qualitative Examples and Analysis}
\label{app:qualitative}

\autoref{fig:arc_facts} and \autoref{fig:medqa_facts} list the top-10 most frequent facts used for training questions, as well as a sample of 10 from the long tail of facts used for only one question. We also show which of these facts are retained in the partial coverage 100-$\text{Mt}_{PC}$.
The top ARC facts are core pieces of knowledge very similar to those that appear in the hand-crafted WorldTree corpus; many of the facts in MedQA are about chronic obstructive pulmonary disease (COPD), reflecting that the Smoking topic is the largest in the training set and that COPD is a common medical condition associated with smoking.

\begin{figure}[t!]
    \centering
    \begingroup
    \renewcommand{\arraystretch}{1.2}
    
    \scriptsize
\begin{tabular}{p{11.3cm}cc}
\toprule
Fact & \# Qs & Mt$_{\text{PC}}$ \\
\midrule
any feature which is determined by genes that can be transmitted from parent to offspring during reproduction is considered an inherited trait & 23 & \up Yes \\
the earth turns on its axis from west to east once every 24 hours & 22 & \up Yes \\
sedimentary rocks are most often formed by the compaction and cementation of sediment, which is material transported and deposited by wind, water, or ice & 20 & \up Yes \\
the tilt of earth's axis is the reason for the changing amount of sunlight that each hemisphere receives, which in turn is the driver for seasons & 20 & \up Yes \\
genes are the instructions for physical traits that pass from parents to offspring & 20 & \up Yes \\
Though Earth’s revolution around the Sun does affect the duration of daylight and darkness, the basic existence of the two phenomena is due to Earth’s rotation on its axis & 19 & \up Yes \\
inheriting means receiving genetic information and traits from a parent or parents & 19 & \dn No \\
Physical traits of an organism, such as fur color, eye color, and size, can be genetically passed from parents to offspring & 19 & \up Yes \\
The rotation of the Earth regulates the daily cycle of day and night, not the seasonal cycle & 19 & \up Yes \\
the rotation of the Earth on its axis is the movement of the Earth around a central point, once every 24 hours & 18 & \dn No \\
\midrule
Whales utilize ocean currents and temperature gradients to guide their migration routes & 1 & \dn No \\
Mutations can be caused by various factors such as exposure to radiation, certain chemicals, or errors during DNA replication & 1 & \dn No \\
a tree requires sunlight to grow & 1 & \dn No \\
lava beds are igneous rock, formed from cooled lava & 1 & \dn No \\
wilting is a survival mechanism, limiting water loss by reducing the surface area exposed to the sun & 1 & \dn No \\
Mendel's experiments with pea plants were focused on heredity & 1 & \dn No \\
\bottomrule
\end{tabular}
\endgroup
    \caption{Top-10 most used and random 10 least used (at least once) facts in the ARC microtheory. The rightmost column depicts whether the fact is retained in the ``partial coverage'' microtheory (Mt$_{\text{PC}}$) that reduces redundancy while maximizing training coverage.}
    \label{fig:arc_facts}
\end{figure}
\begin{figure}[t!]
    \centering
    \begingroup
    \renewcommand{\arraystretch}{1.2}
    
    \scriptsize
\begin{tabular}{p{10.3cm}cc}
\toprule
Fact & \# Questions & Mt$_{\text{PC}}$ \\
\midrule
The severe smoker with an estimated 40 pack-year history has a high risk for chronic obstructive pulmonary disease (COPD) and lung cancer & 45 & \up Yes \\
The man's history of long-term smoking puts him at increased risk for cardiovascular and lung diseases, including pulmonary embolism & 34 & \up Yes \\
a heavy smoker with a 25-pack-years history is at a high risk for lung diseases & 32 & \dn No \\
Chronic obstructive pulmonary disease (COPD) is generally seen with a history of smoking and presents with symptoms such as shortness of breath, progressive dyspnea, cough, and sputum production & 31 & \up Yes \\
a 50 pack-year smoking history contributes to a multitude of health risks, including lung cancer & 27 & \dn No \\
findings such as blood and protein in the urine, as well as high serum urea nitrogen and creatinine levels, indicate kidney damage & 25 & \up Yes \\
A 40 pack-year smoking history, hypertension and diabetes make one more likely to have COPD & 25 & \dn No \\
a man with a 50 pack-year smoking history has a very high risk of developing health issues related to smoking & 25 & \dn No \\
long-term exposure to lung irritants, particularly tobacco smoke, is the primary risk factor for developing chronic obstructive pulmonary disease & 24 & \dn No \\
A patient's history of smoking cigarettes daily for 25 years can have a significant impact on their respiratory function, leading to problems such as shortness of breath, especially after a surgical procedure & 22 & \dn No \\
\midrule
A rapidly enlarging scalp lesion in a child, especially if blanching with pressure, is suggestive of a vascular lesion such as a strawberry hemangioma & 1 & \dn No \\
losing a job due to issues such as tardiness is likely due to some negative circumstance impacting the individual's life, such as alcohol abuse & 1 & \dn No \\
Bloody diarrhea can be a symptom of amoebic dysentery caused by amoebic infection & 1 & \dn No \\
fever is a common symptom in vasculitis because of the inflammation & 1 & \dn No \\
Normal-pressure hydrocephalus is characterized by cognitive changes, gait problems, and urinary incontinence & 1 & \dn No \\
Blood pressure of 197/124 mm Hg is considered dangerously high and needs to be lowered immediately & 1 & \dn No \\
Creatinine phosphokinase is an enzyme found in the heart, brain, skeletal muscle and other tissues, its levels increase when muscle or heart cells are injured & 1 & \dn No \\
the liver breaks down bilirubin so it can be removed from the body in the stool & 1 & \dn No \\
Smoking and unprotected sex increase the risk of various infections, including urinary tract infections & 1 & \dn No \\
Regular kneeling can put excessive pressure on prepatellar bursa, potentially causing inflammation and pain & 1 & \dn No \\
\bottomrule
\end{tabular}

\endgroup
    \caption{Top-10 most used and random 10 least used (at least once) facts in the MedQA microtheory. The rightmost column depicts whether the fact is retained in the ``partial coverage'' microtheory (Mt$_{\text{PC}}$) that reduces redundancy while maximizing training coverage.}
    \label{fig:medqa_facts}
\end{figure}

\autoref{fig:arc_mt} depicts a subgraph of a much larger network of entailments connecting statements in $\mathcal{C}$ to 18 of the training hypotheses. It shows which of these statements were retained by the partial coverage Mt for their support of multiple related hypotheses. The subgraph illustrates 4 cliques; one for hypotheses about the gravitational force between opbjects, and 3 about the force, acceleration and mass of objects in motion. The core statements retained in the Mt are mainly versions of Newton's second law of physics, $F = m \cdot a$ and Newton's Law of universal gravitation. While a singular fact describing the law might theoretically serve the same purpose in proving all these hypotheses, We observe that the entailment engine prefers to use separate facts representing different interpretations of the law (e.g. how to calculate net force, vs how to calculate the force required to move an object).
This may also be a result of imposing a depth-2 limit of the training proof search. It would perhaps require more ``legwork'' by the entailment engine to construct a chain of more than 2 entailments from a single statement, e.g. just `force equals mass times acceleration,' to each hypothesis.

\begin{figure}[H]
    \centering
    \includegraphics[trim={12mm 0 12mm 0},clip,width=\textwidth]{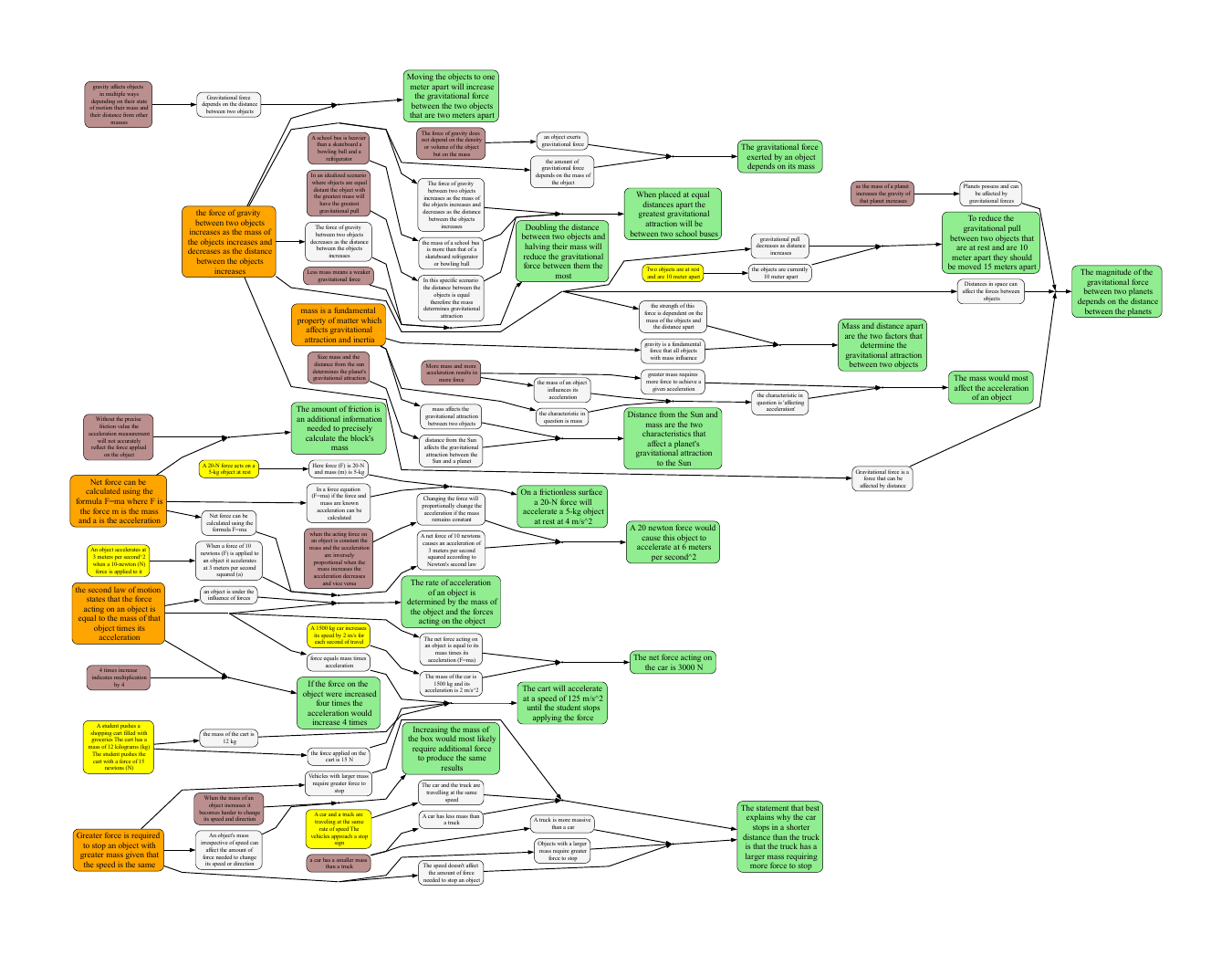}
    \caption[Example subgraph show how ARC microtheory statements are used to ground hypotheses.]{Subgraph showing some of the statements from the 200-$\text{Mt}_{\text{PC}}$ ARC microtheory (\highlighthtml{FFA500}{orange}) and their influence on proving training hypotheses (\highlighthtml{90EE90}{green}) in different question contexts (\highlighthtml{ffff00}{yellow}). The partial coverage (PC) algorithm selects facts from $\mathcal{C}$ that provide coverage for multiple questions. Facts from $\mathcal{C}$ that are specific to only one or two questions (\highlighthtml{BC8F8F}{maroon}) are not selected. Each dot ($\cdot$) represents an entailment; inferences (\highlighthtml{f5f5f5}{grey} and \highlighthtml{90EE90}{green}) can be supported by alternative entailments in the form of multiple entering edges.
    In this instance, the PC algorithm shows to retain 5 facts mostly related to Newton's second law, which is core to multiple subtopics in grade school physics tests (motion, friction, gravitational attraction). Only the \highlighthtml{FFA500}{orange} nodes are retained in the microtheory for test time inference; all other statements
    are discarded.
    }
    \label{fig:arc_mt}
\end{figure}

\autoref{fig:medqa_mt} shows a similar subgraph for the MedQA microtheory, this time illustrating the facts retained by the full question coverage 100-$\text{Mt}_{\text{QC}}$. Hypotheses are fully thus grounded to Mt statements and the context passages of their respective questions. This graph also illustrates the effect of retaining inferences during the serial computation of the $L_{kj}$ leaf sets.
\begin{figure}[H]
    \centering
    \includegraphics[trim={12mm 0 12mm 0},clip,width=\textwidth]{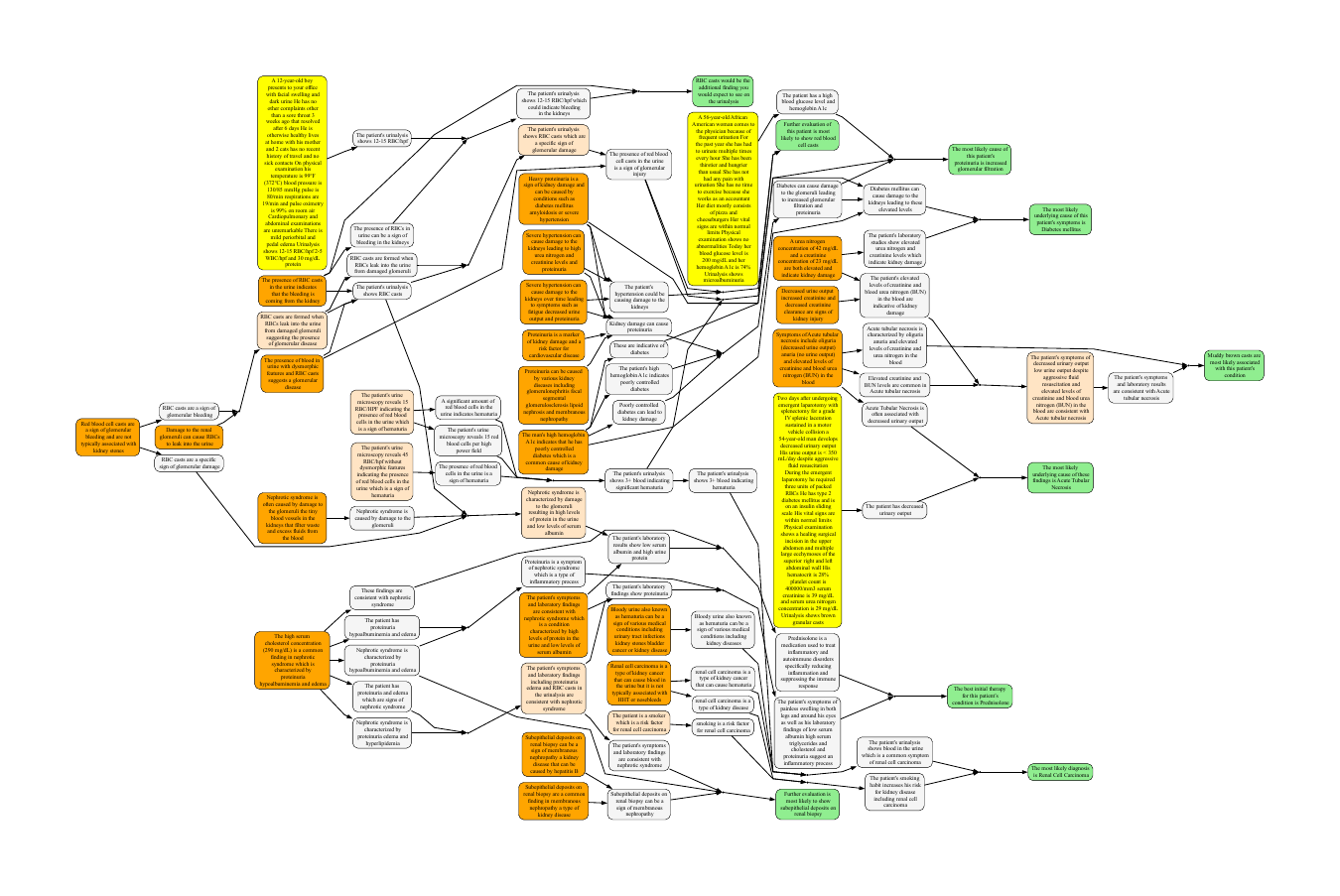}
    \caption[Example subgraph show how MedQA microtheory statements are used to ground hypotheses.]{Subgraph showing some of the statements from the 100-$\text{Mt}_{\text{QC}}$ MedQA microtheory (\highlighthtml{FFA500}{orange}) and their influence on proving training hypotheses (\highlighthtml{90EE90}{green}) in different question contexts (\highlighthtml{ffff00}{yellow}) related to clinical cases.
    Unlike the PC approach, the question coverage (QC) approach ensures full grounding of each training hypothesis using facts from the microthery.
    As MedQA hypotheses often require many entailment hops, the inference retention mechanism described in \S\ref{sec:retaining} helps prove hypotheses using inferences (\highlighthtml{FFE4C4}{light orange}) grounded from previous questions, some of which are omitted for space.
    }
    \label{fig:medqa_mt}
\end{figure}

\section{Retrieval Implementation Details}
The Wikipedia corpus used in \S\ref{sec:engine_exps} is structured as a flat index (i.e., a very long list) of paragraph chunks– specifically, the  2021-01-
20 version of Wikipedia with 100 word chunks. For similarity, we split the textbooks into 100 word chunks as well.
We retrieve paragraphs using two-phase retrieval; we use BM25~\cite{robertson-etal-1995-okapi} for first stage retrieval and rerank using SentenceTransformer’s ms-marco-MiniLM-L-12-v2~\cite{reimers-gurevych-2019-sentence}. 

When we retrieve facts from both the Microtheory and Wikipedia (or Wikipedia+Textbooks), we append the sets in that order. When we retrieve from Wikipedia+Textbooks, we interleave the results one-by-one to give equal precedence to the two sources. These design choices were chosen based on empirical performance.

\section{Mixtral Results on MedQA}
\label{app:mixtral-results}
To demonstrate our work is not closed-model dependent, this section replicates the MedQA experiments from the paper using 
Mixtral-8x22B-Instruct-v0.1~\cite{jiang-etal-2024-mixtral} as the source of the microtheories instead of GPT-4. 
\autoref{tab:stats_mistral}, 
\autoref{fig:fact_usage_hist_mistral},
\autoref{fig:test_coverage_mixtral},
and 
\autoref{fig:relevance_expert_results_mixtral}
each show Mixtral results analogous to an experimental result from the main paper using GPT-4. We have commented on the similarities and differences with the original result in the caption of each figure.

\begin{figure}[t!]
    \centering
    \begin{minipage}[t]{0.52\textwidth}
        \centering
        \adjustbox{valign=c} {\footnotesize
        \begin{tabular}{@{}lc@{}}
        \toprule 
         \textbf{Mixtral-8x22b Results} &  \textbf{MedQA} \\ \midrule
         \multicolumn{2}{@{}l}{\textbf{Dataset Details}} \\
        \# Questions      & 641/94/186 \\
        \# Topic Clusters      & 4 \\
        Ext. Corpus  & Wiki+Textbooks \\
        \midrule 
        \multicolumn{2}{@{}l}{\textbf{Training Extraction Results}} \\
        $|\mathcal{F}|$  & 11,082 \\
        $|\mathcal{C}|$  &  8,183  \\ 
        {\# Qs with Proof}           & 476 (74\%) \\
        {Min \#Fs to Cover}         & 918          \\
        {Fact/Q Ratio}                      & 1.91         \\ \bottomrule
        \end{tabular}}
        \caption{Results of microtheory extraction from training data using Mixtral-8x22b as the underlying LLM. The last two rows result from the ``min-fact'' LP.
        Compared with GPT-4, Mixtral generates 5000 fewer unique facts (11K vs 16K) and finds proofs for 14\% fewer questions. 
        }
        \label{tab:stats_mistral}
    \end{minipage}%
    \hfill
    \begin{minipage}[t]{0.45\textwidth}
        \centering
        \adjustbox{valign=c}{\includegraphics[trim={0 3mm 0 3.6cm},clip,width=\linewidth]{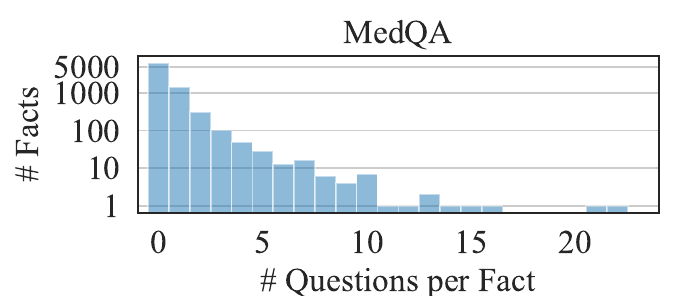}}
        \caption{Histogram of training question per fact in $\mathcal{C}$
        and 650 MedQA training questions using Mixtral-8x22b as the underlying LLM.
        While multiple GPT-4 generated facts are used for 25 or more training questions, no Mixtral-generated fact is used more than 22 times.
        }
        \label{fig:fact_usage_hist_mistral}
    \end{minipage} 
    
    \begin{minipage}[t]{0.52\textwidth}
        \centering
        \adjustbox{valign=c}{\begin{tabular}[t]{@{}c@{}}
            \includegraphics[trim={0 0 0 4pt},clip,width=\linewidth]{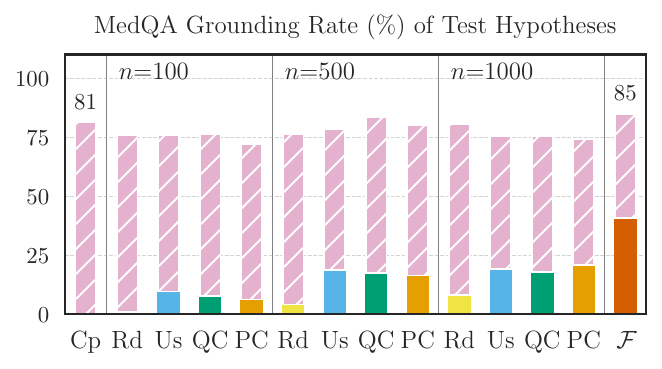}
        \end{tabular}}
        \vspace{-3mm}
        \caption{
        \% of test hypotheses fully groundable after adding various Mixtral-generated microtheories ($n$-sized \textbf{R}an\textbf{d}om, \textbf{Us}age, \textbf{QC}, \& \textbf{PC} Mts) to a base \textbf{C}or\textbf{p}us.
        We show the fraction of leaves grounded in the Mt (solid) and the base corpus (striped). 
        }
        \label{fig:test_coverage_mixtral}
    \end{minipage}%
    \hfill
    \begin{minipage}[t]{0.45\textwidth}
        \centering
        \adjustbox{valign=c}{\includegraphics[width=\linewidth]{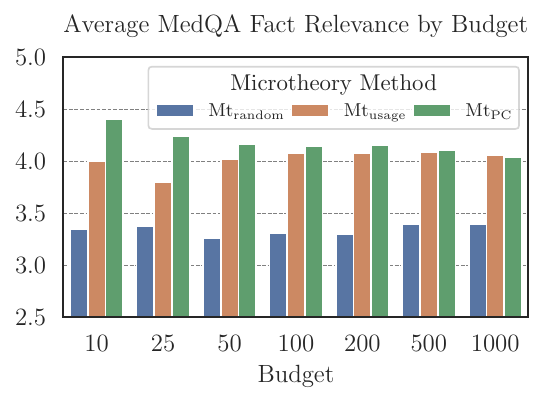}}
        \vspace{-4mm}
        \caption{Average expert-annotated task relevance score for Mixtral-generated microtheory facts.
        }
    \label{fig:relevance_expert_results_mixtral}
    \end{minipage}
\end{figure}

\section{Additional Fact Usage Metrics}
\label{app:fraction_usage}
In addition to the grounding rate (\sref{sec:test_grounding}), human-annotated relevance rate (\sref{sec:human_relevance}), and automated relevance rate (\sref{sec:auto-relevance}), this section shows a pair of additional metrics for our considered microtheories:
(1) the fraction of all facts in a microtheory used by the entailment engine for at least one test question, and (2) the average number of test questions for which each fact in the microtheory is used.

\autoref{fig:fraction_used} shows the fraction of microtheories used at least once at test time, and \autoref{fig:avg_usage} shows the average number of questions for which any single Mt fact is used. 
In both cases, the $\text{Mt}_{\text{usage}}$,$\text{Mt}_{\text{QC}}$, and $\text{Mt}_{\text{PC}}$ have higher usage than the random sample baseline, while the full fact pool $\mathcal{F}$ scores very low, since most of the facts in the full pool are never used at test time. 
Expectedly, the $\text{Mt}_{\text{usage}}$ approach scores highest on both metrics, given this is what it was optimized for; we sorted facts by their usage counts at training time (the same metric measured in \autoref{fig:avg_usage}) and took the top-$n$ as the Mt. 
We note that this does not directly correlate with Mt quality since, as we discussed in \sref{sec:usage_optimization}, sorting only by per-fact usage fails to remove functional redundancies. The test grounding rates in \autoref{sec:test_grounding} should be considered a more encompassing metric.

\begin{figure}[t!]
    \centering
    \begin{minipage}[t]{0.49\textwidth}
        \centering
        \includegraphics[trim={0cm 3mm 0cm 0cm},clip,width=\linewidth]{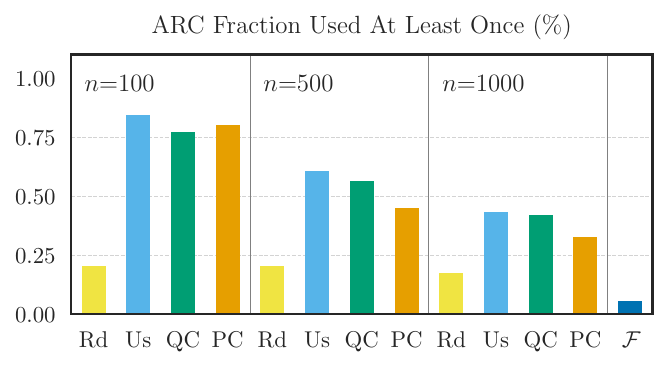}\\[\baselineskip]
        \includegraphics[trim={0cm 0mm 0cm 2mm},clip,width=1\linewidth]{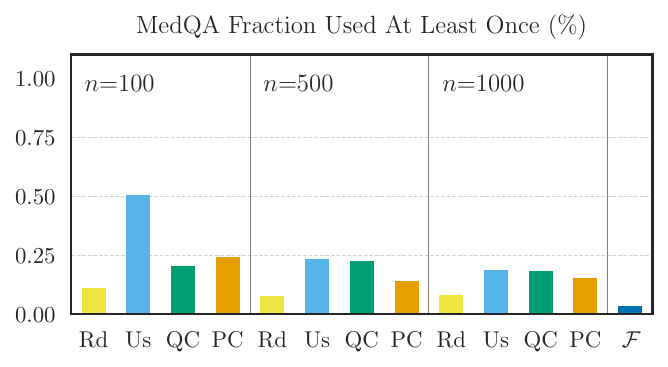}
        \caption{Fraction of a microtheory used in an entailment proof for at least one test question (out of 249 for ARC, 186 for MedQA). All distillation methods ($n$-sized \textbf{Us}age, \textbf{QC}, \& \textbf{PC} Mts) are used substantially more frequently than the random sample baseline, with the \textbf{Us}age having the highest fraction. The fraction used for the full raw fact pool $\mathcal{F}$ is very low.
        }
        \label{fig:fraction_used}
    \end{minipage} 
    \hfill
    \begin{minipage}[t]{0.49\textwidth}
        \centering
        \includegraphics[trim={0cm 3mm 0cm 0cm},clip,width=1\linewidth]{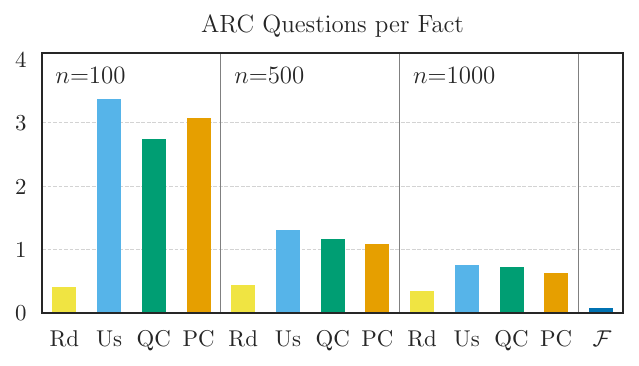}\\[\baselineskip]
        \includegraphics[trim={0cm 0mm 0cm 2mm},clip,width=\linewidth]{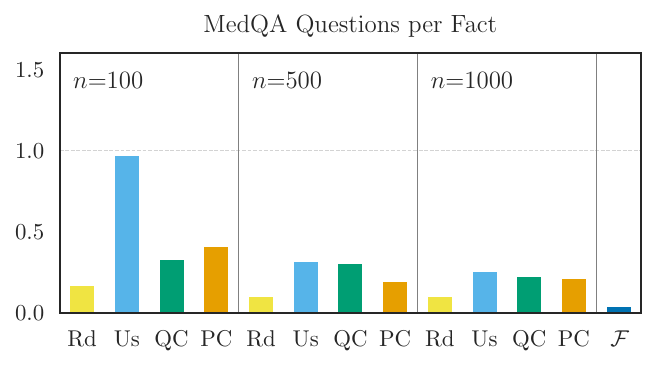}
        \caption{Average number of test questions (out of 249 for ARC, 186 for MedQA) per fact in a microtheory. Distilled microtheories are used in more questions than the random baseline, with the average usage of \textbf{Us}age microtheory facts being (expectedly) highest.  
        }
        \label{fig:avg_usage}
    \end{minipage}
\end{figure}

\section{Extended Relevance Rate Results}
\autoref{fig:relevance_results} shows the rate at which the considered microtheory methods contain 5-rated relevant facts for the test sets in the topic mini-splits of ARC and MedQA.
\begin{figure}[t!]
    \centering
    \includegraphics[trim={2mm 1cm 2mm 1.4cm},clip,width=1\linewidth]{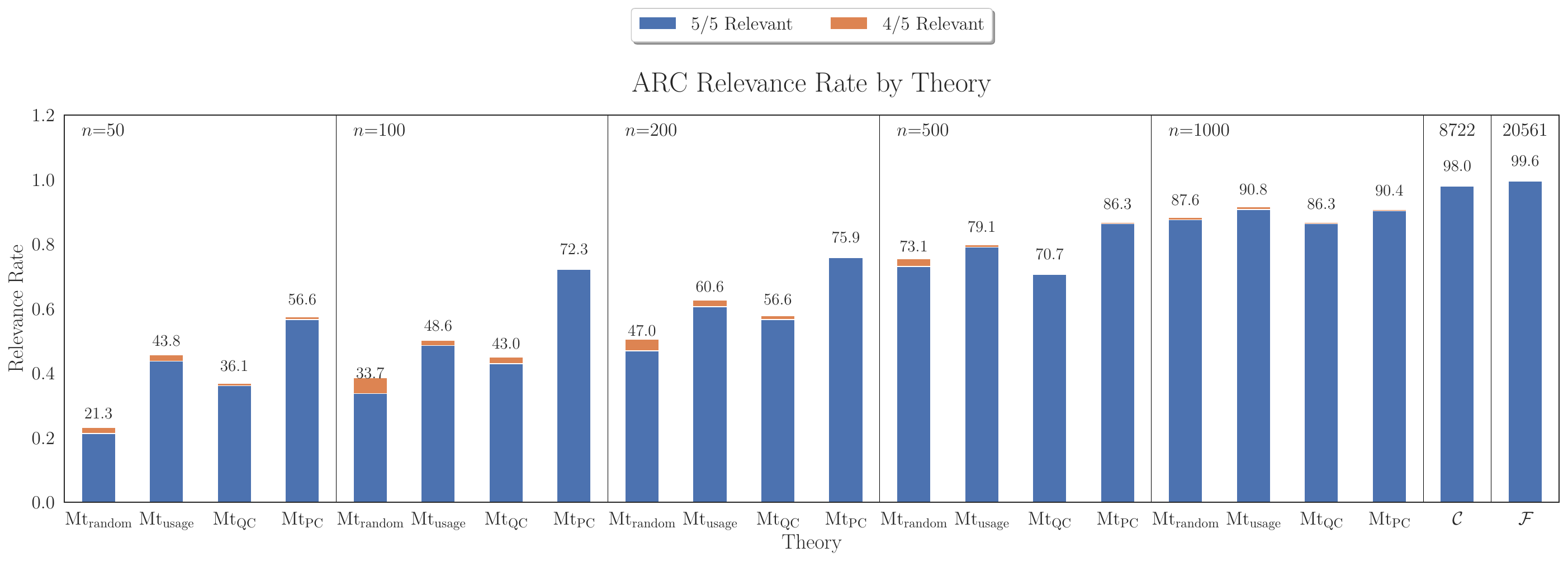} \\
    \includegraphics[trim={2mm 0cm 2mm 0cm},clip,width=\linewidth]{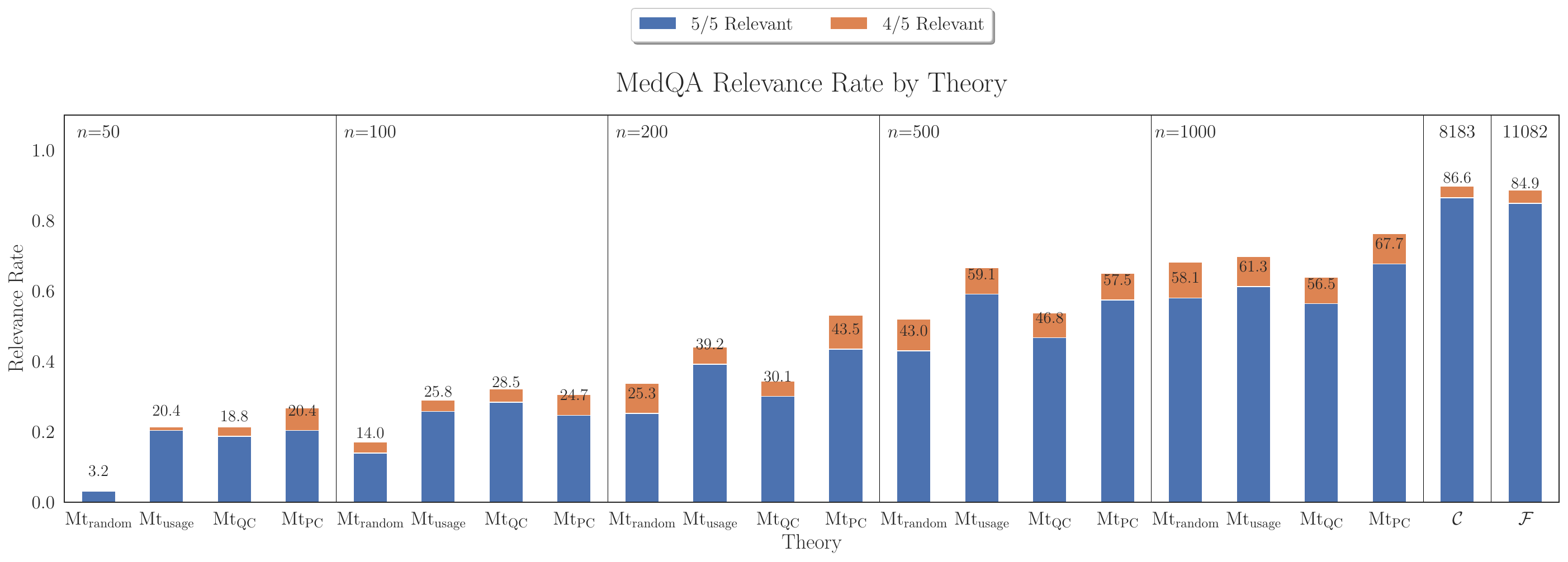}
    \caption{Rate (\%) of microtheories containing at least one fact with 5/5 and 4/5 relevance as determined by GPT-4o. Numbers above each bar are the percentage of 5's. }
    \label{fig:relevance_results}
\end{figure}

\section{Expert Relevance Annotation Details}
\label{app:expert-ratings}

We sampled 25 facts (fewer when necessary) from Mts with 70 budgets of size ranging from 10 to 1000. We evaluated the Mt$_\text{usage}$, Mt$_\text{PC}$, and Mt$_\text{random}$ approaches. Both annotators rated each fact independently and then reconciled the (rare) disagreements.
We computed agreement metrics across the full set of 314\footnote{This number is not 7 $\times$ 3 $\times$ 25 because some Mts were smaller than 25 facts, the $n$-Mt$_\text{usage}$s are all supersets of each other, and many other samples contained high overlap due to the extraction algorithm behavior.} annotated facts, finding a Krippendorff $\alpha$ of .81, Cohen's $\kappa$ of .809, and raw agreement rate of .863.

\begin{figure}[t!]
    \centering
    \includegraphics[trim={2.5cm 2.5cm 1.5cm 3cm},clip,width=\linewidth]{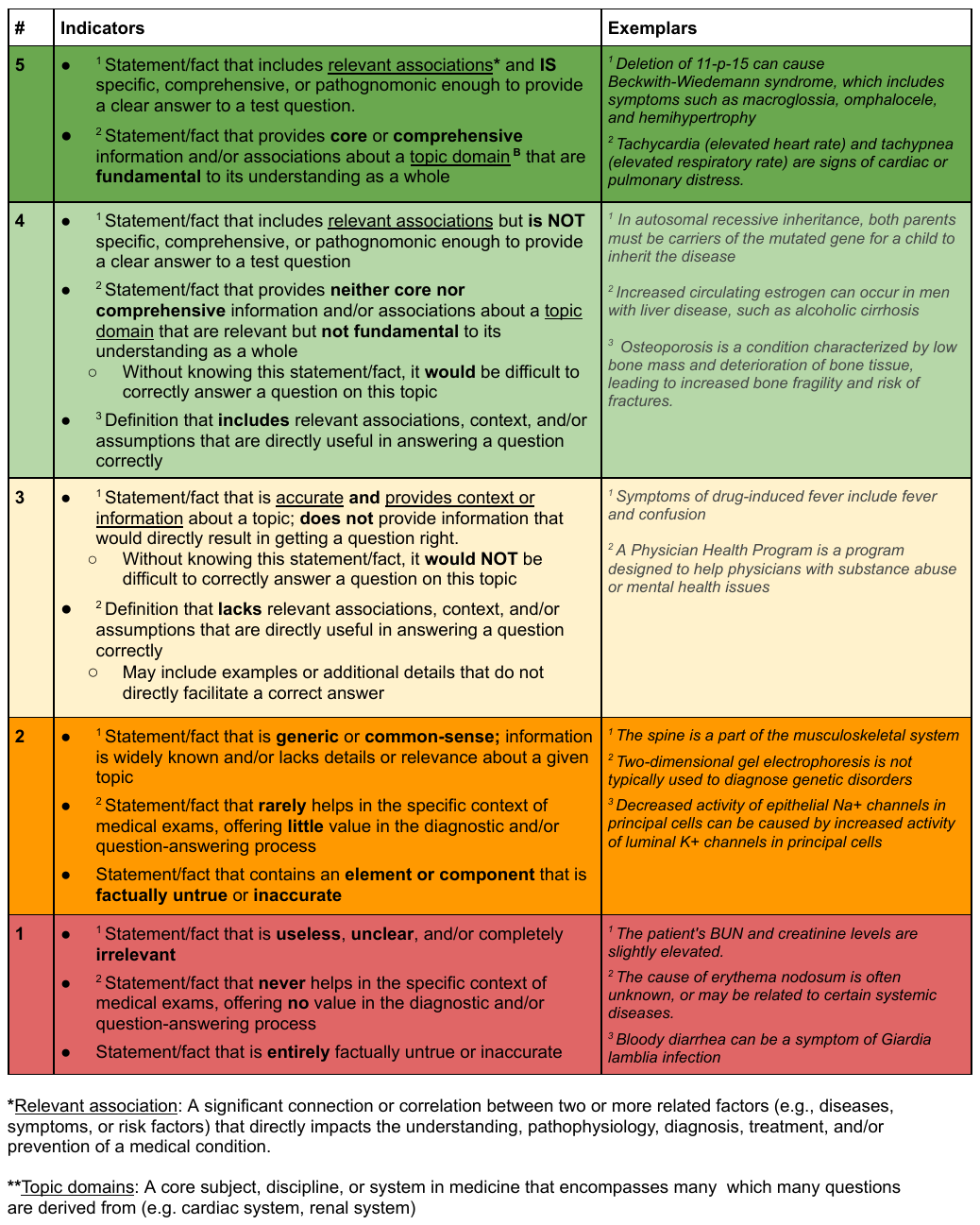}
    \caption{Rubric used to collect expert relevance ratings for medical microtheory facts (1 of 2)}
    \label{fig:medqa_rubric_1}
\end{figure}

\begin{figure}[t!]
    \centering
    \includegraphics[trim={2.5cm 12.5cm 1.5cm 3cm},clip,width=\linewidth]{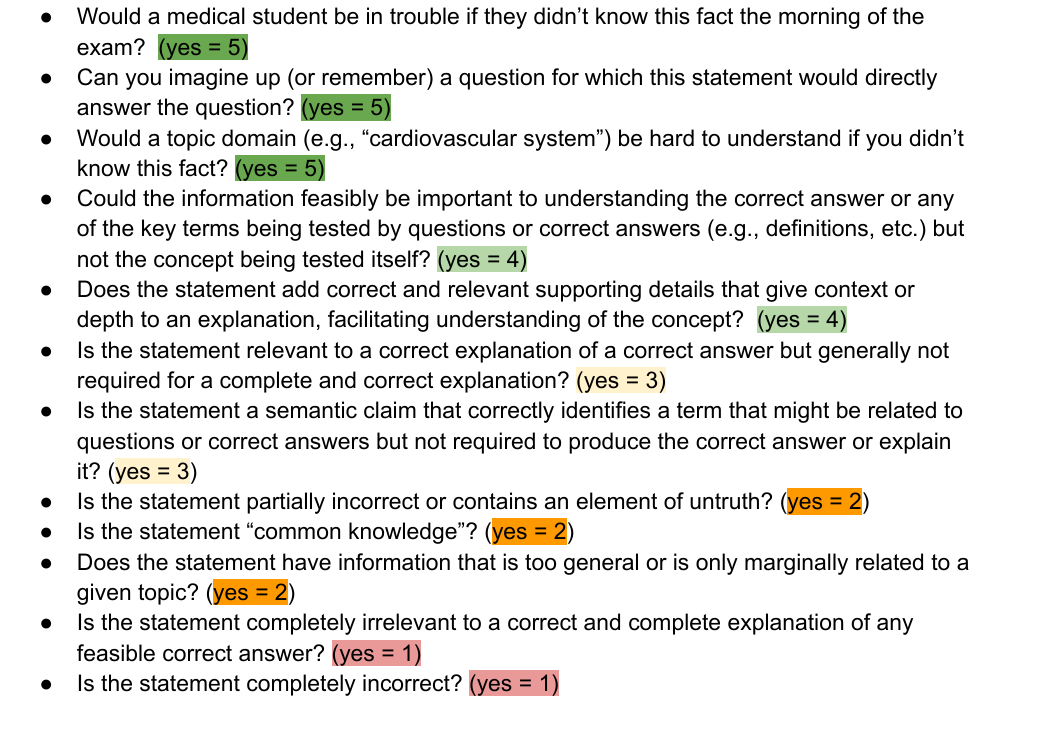}
    \caption{Rubric used to collect expert relevance ratings for medical microtheory facts (2 of 2)}
    \label{fig:medqa_rubric_2}
\end{figure}

\section{Automatic Relevance Annotation Details}
\label{app:auto_relevance_prompt}
The prompt used for scoring the relevance of facts to a given question is shown in \autoref{fig:relevance-prompt}. The original rubric by \citet{jansen-etal-2021-challenges} contains 4 ordinal scores; we added a 5th to differentiate between facts relevant to ruling out incorrect options (4) vs supporting correct options (5).

\begin{figure}[t!]
    \centering
    \scriptsize
    \begin{lstlisting}[style=base,basicstyle=\tiny\ttfamily]
You are a scientific expert that helps other scientists determine whether various facts are relevant to explaining the correct answer to a question. Scientists give you a question and a list of facts that they believe are relevant to the question. Your job is to determine whether each fact is relevant to explaining the correct answer to the question using a provided rubric for scoring:


A fact has a score of 5 if it has the following indicators:
* core/critical fact/knowledge that explains why a correct answer is correct
* essential piece of information if explaining phenomenon in question to a toddler

A fact has a score of 4 if it has the following indicators:
* core/critical fact/knowledge that explains why an incorrect answer is incorrect

A fact has a score of 3 if it has the following indicators:
* moderately important fact necessary in a correct explanation but not being explicitly tested *underlying assumption necessary to understanding the terms used in the correct response and explanation
* defines terms used in the prompt and/or the correct answer
* semantic statements: correctly identify a term in the immediate question or correct answer
* common sense statement that is topically relevant but not necessary for differentiating correct from incorrect 

A fact has a score of 2 if it has the following indicators:
* extra detail - explanations missing this fact are neither incorrect nor incomplete
* true relevant statement but not necessary for understanding phenomena in question

A fact has a score of 1 if it has the following indicators:
* irrelevant 
* incorrect


You do NOT need to give a 5 for at least one fact. You should give a 5 only if the fact is absolutely necessary and core to correctly answer the immediate question (i.e. if someone does not know the fact, they cannot answer the question correctly). 

Here are some of the questions that you ask yourself when considering a fact:
Does this statement directly answer the original question? (yes = 5)
* Is the information absolutely necessary to correctly answer the immediate question? (yes = 5)
* Is the information absolutely necessary to explain why another answer is incorrect for the immediate question? (yes = 4)
* Is the information important to understanding the correct answer or any of the key terms being tested by the immediate question or correct answer (e.g., definitions, etc.) but not the concept being tested itself? (yes = 3)
* Is the statement a semantic claim that correctly identifies a key term in the immediate question or correct answer in a way that is relevant to the phenomenon being tested? (yes = 3)
* Is the statement relevant to a correct explanation of the correct answer but not required for a complete and correct explanation? (yes = 2)
* Does the statement add correct and relevant supporting details that give context or depth to an explanation but are not required to make the explanation complete and correct? (yes = 2)
* Is the statement a semantic claim that correctly identifies a term that is related to the immediate question or correct answer but not required to produce the correct answer or explain it? (yes = 2)
* Is the statement completely irrelevant to a correct and complete explanation of the correct answer? (yes = 1)
* Is the statement incorrect? (yes = 1)
    \end{lstlisting}
    \caption[Prompt used to score the question relevance of retrieved microtheory statements.]{Prompt used to score the question relevance of retrieved microtheory statements. Instructions are based on the rubric from \citet{jansen-etal-2021-challenges}. }
    \label{fig:relevance-prompt}
\end{figure}

\end{document}